\documentclass[11pt]{article}

\usepackage[final]{acl}

\usepackage{times}
\usepackage{latexsym}
\usepackage{afterpage}
\usepackage{stfloats}
\usepackage[T1]{fontenc}

\usepackage[utf8]{inputenc}

\usepackage{microtype}

\usepackage{inconsolata}

\usepackage{graphicx}
\usepackage{multirow}
\usepackage{tabularx}
\usepackage{amsmath}
\usepackage{graphicx}
\usepackage{subfig}
\usepackage{soul}
\usepackage{enumitem}
\usepackage{multirow} 
\usepackage{booktabs} 
\usepackage[most]{tcolorbox}
\usepackage{listings}
\usepackage{xcolor}
\usepackage{caption} 
\usepackage{subfig}
\usepackage{xcolor}
\usepackage{amssymb}

\lstdefinestyle{eventstyle}{
    basicstyle=\ttfamily\scriptsize,
    breaklines=true,
    breakatwhitespace=true,
    postbreak=\mbox{\textcolor{gray}{$\hookrightarrow$}\space},
    columns=fullflexible,
    keepspaces=true,
    showstringspaces=false,
    xleftmargin=2pt,
    xrightmargin=2pt,
    aboveskip=2pt,
    belowskip=2pt
}

\newtcolorbox{eventbox}[1]{
    enhanced,
    colback=blue!3,
    colframe=blue!75!black,
    coltitle=white,
    colbacktitle=blue!75!black,
    title={#1},
    fonttitle=\bfseries\small,
    boxrule=0.8pt,
    arc=2pt,
    left=5pt,
    right=5pt,
    top=5pt,
    bottom=5pt,
    attach boxed title to top left={
        xshift=0pt,
        yshift=-1pt
    },
    boxed title style={
        sharp corners,
        boxrule=0pt,
        colback=blue!75!black,
        left=6pt,
        right=6pt,
        top=2pt,
        bottom=2pt
    }
}
\title{Denoising the Future: Context-Aware Spectral Diffusion\\ for Temporal Knowledge Graph Extrapolation}

\author{
 \textbf{Yanglei Gan\textsuperscript{1}},
 \textbf{Peng He\textsuperscript{2,4}},
 \textbf{Run Lin\textsuperscript{3}},
 \textbf{Peiyuan Jiang\textsuperscript{2}},
 \textbf{Yifan Wang\textsuperscript{2}},
 \textbf{Qiao Liu\textsuperscript{2}}
\\
 \textsuperscript{1}Southwest Minzu University, Chengdu, China\\
 \textsuperscript{2}University of Electronic Science and Technology of China, Chengdu, China\\
 \textsuperscript{3}Zhejiang University, Hangzhou, China, 
 \textsuperscript{4}Weixin Group, Tencent, Gunagzhou, China
\\
 \small{
   \textbf{Correspondence:} \href{mailto:email@domain}{emmaahe@tencent.com, runlin@zju.edu.cn}
 }
}

\begin{document}
\maketitle
\begin{abstract}


Temporal Knowledge Graph (TKG) extrapolation seeks to infer future facts from time-varying relational histories. Recent diffusion-based approaches improve uncertainty modeling through generative denoising, but their aggregated conditioning on subject histories may insufficiently distinguish query-specific evidence from non-salient historical facts, thereby diluting target-discriminative signals. To bridge this gap, we propose \textbf{FreqDiff}, a \textbf{Freq}uency-aware \textbf{Diff}usion framework for TKG extrapolation. Specifically, FreqDiff formulates future object prediction as query-slot denoising and develops a dual-stream denoiser that integrates temporal dependency modeling with context-aware spectral calibration. The spectral branch synthesizes history-conditioned filters from learnable bases to adaptively re-calibrate denoising representations, while a frequency-domain regularizer is proposed to align the denoised target with the gold object in spectral space. Experiments on four public TKG benchmarks demonstrate that FreqDiff achieves state-of-the-art performance\footnote{The source code is available at: \url{https://github.com/AONE-NLP/FreqDiff}.}.

\end{abstract}

\section{Introduction}
Temporal Knowledge Graphs (TKGs) encode time-evolving facts as quadruples $(s, r, o, t)$, where relation $r$ links entities $s$ and $o$ at timestamp $t$ \cite{ji2021survey,liang2024survey}. Reasoning over TKGs aims to infer missing or future facts from observed temporal histories. Existing studies typically distinguish between two reasoning settings: \textbf{interpolation}, which completes missing facts within the observed time span \cite{cai2023temporal,luo2024chain}, and \textbf{extrapolation}, which predicts facts after the latest observed timestamp \cite{yang2015embedding,zhu2021learning,Li21Temporal}. This work centers on extrapolation, which enables forward-looking reasoning and supports decisions about future events.

\begin{figure}[t]
    \centering
    \includegraphics[width=\linewidth]{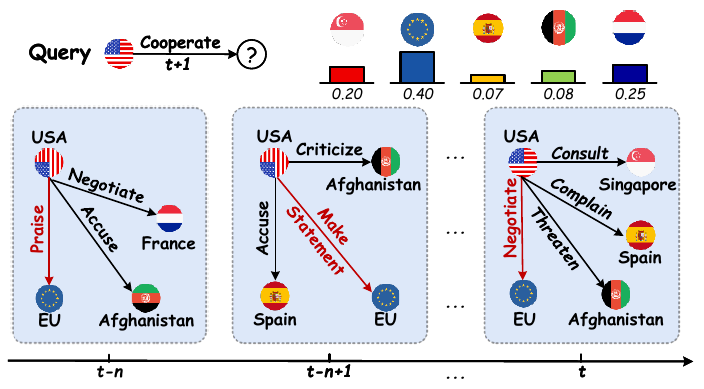}
    \caption{Illustration of subject-oriented context in TKG. \textcolor{red}{Red}-linked facts denote query-oriented evidence, while other subject-related facts may act as contextual noise.}
    \label{intro}
\end{figure}

Accurate TKG extrapolation requires modeling the temporal evolution of events and relational patterns. Most existing methods follow a learn-to-classify paradigm \cite{trivedi2017know,jin2020recurrent}, which encodes historical dependencies into deterministic entity and relation representations and ranks candidate future facts with scoring functions such as TransE \cite{bordes2013translating} or DistMult \cite{yang2015embedding}. Recent work further improves this paradigm through GNN-based historical propagation \cite{Li21Temporal,li2022tirgn}, contrastive learning \cite{chen2024local,xu2023temporal} over local/global or historical/non-historical contexts, and symbolic temporal priors \cite{chen2025enhancing,chen2025cogntke} for interpretable reasoning.



Although these methods achieve strong empirical performance, their deterministic prediction mechanism makes it difficult to capture the uncertainty and diversity of future events. To address this limitation, recent studies have introduced diffusion models into TKG reasoning and shifted toward a \textbf{learn-to-generate} paradigm \cite{cai-etal-2024-predicting,gan2026negative}, where plausible target objects are generated or sampled conditioned on historical temporal contexts. Despite their empirical success, existing methods exhibit two key limitations: 

\begin{itemize}[leftmargin=*, itemsep=0pt, parsep=0pt] 
    \item \textbf{Limited Discrimination of Query-specific Evidence.} Existing methods typically encode subject-oriented histories through unified temporal processing schemes \cite{zhang2023learn,chen2024natural}. However, a subject history may contain diverse relational trajectories \cite{han2021explainable,wang2025dltkg}, not all of which are informative for the current query. As shown in Figure \ref{intro}, for the query <USA, Cooperate, ?, t+1>, accusation- or criticism-related facts are subject-relevant but weakly query-discriminative, whereas negotiation- or consultation-related facts provide more direct evidence. Without query-specific filtering, such non-salient contexts may obscure truly critical temporal evidence.
    \item \textbf{Reliance on Generic Loss Formulations. } Existing diffusion-based TKG reasoning methods mainly supervise denoising with generic reconstruction or ranking losses \cite{cai-etal-2024-predicting,gan2026negative}. Although effective for entity discrimination, these objectives constrain the denoised representation mostly in the embedding space and do not explicitly preserve its spectral structure. Consequently, frequency-specific cues important for reconstructing the future object may be insufficiently captured.
\end{itemize}

In this paper, we propose \textbf{FreqDiff}, a \textbf{Freq}uency-aware \textbf{Diff}usion framework for TKG extrapolation. Given a future query, FreqDiff constructs a subject-oriented event sequence as historical context and formulates missing object prediction as query-slot denoising. In the forward process, Gaussian noise is injected only into the target object representation, while historical events remain deterministic. In the reverse process, a dual-stream denoiser reconstructs the target by combining temporal dependency modeling with context-aware spectral calibration. The temporal branch captures sequential patterns from the subject history, while the spectral branch generates context-aware spectral filters from learnable bases to re-calibrate the denoising representation. Moreover, a frequency-domain consistency regularizer aligns the denoised target with the gold object embedding in spectral space, providing explicit supervision for frequency-aware reconstruction. Our contributions are three-fold:

\begin{itemize}[leftmargin=*, itemsep=0pt, parsep=0pt, topsep=0pt]
    \item We propose FreqDiff, a frequency-aware diffusion framework for TKG extrapolation, which formulates future object prediction as query-slot denoising and reconstructs the target through a dual-stream denoiser with temporal modeling and context-aware spectral calibration. 
    \item FreqDiff introduces a frequency-domain consistency regularizer that aligns the denoised target representation with the gold object embedding in spectral space, providing explicit supervision for frequency-aware reconstruction.
    \item Extensive experiments on four public TKG benchmarks show that FreqDiff achieves state-of-the-art performance, with further analysis validating the effectiveness of its spectral calibration and frequency-domain regularization.
\end{itemize}

\section{Related Works}

\subsection{Temporal Knowledge Graph Reasoning}

Discriminative TKG reasoning predicts future facts by learning temporal patterns from historical triples. Early continuous-time methods, such as Know-Evolve \cite{bordes2013translating, chen2024thcn} and THCN \cite{chen2024thcn}, model event occurrence with Hawkes processes or temporal causal convolution. Later neural approaches incorporate temporal signals into KG encoders through recurrent reasoning, graph-based propagation, cycle-aware constraints, and structural historical evidence, including RE-NET \cite{jin2020recurrent}, RE-GCN \cite{Li21Temporal}, CyGNet \cite{zhu2021learning}, CEN \cite{li2022complex}, xERTE \cite{han2021explainable}, and HisMatch \cite{li2022hismatch}. 

Beyond simply encoding all historical facts, another line of work explicitly improves the selection or bottlenecking of useful historical evidence. For example, xERTE \cite{han2021explainable} extracts query-relevant temporal subgraphs with temporal relational attention, TimeTraveler \cite{sun2021timetraveler} searches historical snapshots through reinforcement learning, and CENET-style methods distinguish historical and non-historical dependencies through contrastive learning and masking \cite{xu2023temporal,zhang2023learning}. In parallel, symbolic and structural methods improve interpretability and inductive generalization by deriving temporal logical rules from time-consistent random walks \cite{liu2022tlogic}, mining relation-specific paths \cite{dong2023adaptive}, or constructing cognitive temporal relation graphs \cite{chen2025cogntke}.

\subsection{Generative TKG Reasoning}

Generative approaches introduce uncertainty-aware modeling into TKG extrapolation. DiffuTKG \cite{cai-etal-2024-predicting} formulates future fact prediction as conditional denoising under a Gaussian diffusion process, while NADEx \cite{gan2026negative} further incorporates negative-aware diffusion to sharpen decision boundaries for future entities. DPCL-Diff \cite{cao2025dpcl} extends this line with graph-node diffusion and dual-domain periodic contrastive learning to separate recurrent and novel temporal patterns. Beyond purely neural diffusion frameworks, Luo et al. \cite{luo2024chain} use LLMs to generate multi-step event chains, and LLM-DR \cite{chen2025llm} combines classifier-free guided diffusion with LLM-based rule refinement. Despite their progress in generative TKG reasoning, they mainly operate in the time-domain, while the frequency properties underlying temporal dynamics remain under-explored. This motivates our frequency-aware diffusion framework, which introduces complementary spectral signals for TKG reasoning. Discussions of diffusion model and frequency modeling are provided in Appendix \ref{sec:a}.

\section{Preliminary}

\textbf{Definition 1. Temporal Knowledge Graph. } Let $\mathcal{E}$, $\mathcal{R}$, and $\mathcal{T}$ denote finite sets of entities, relation types, and timestamps, respectively. A temporal knowledge graph $\mathcal{G}$ is a collection of time-stamped quadruples:
\begin{equation}
\footnotesize
\mathcal{G}
   \;=\;
   \Bigl\{\, (s,r,o,t) \;\Big|\;
  s,o \in \mathcal{E},\;
  r \in \mathcal{R},\;
  t \in \mathcal{T}
   \Bigr\},
\label{eq:tkg_definition}
\end{equation}
where each tuple encodes the fact that relation $r$ holds from subject $s$ to object $o$ at time $t$. More specifically, the TKG can be viewed as an ordered sequence of static snapshots:
\begin{equation}
\footnotesize
\mathcal{G}
   \;=\;
   \bigl\{
 \mathcal{G}_{1},
 \mathcal{G}_{2},
 \ldots,
 \mathcal{G}_{|\mathcal{T}|}
   \bigr\},
\label{eq:snapshot_sequence}
\end{equation}
where $\mathcal{G}_{t}$ aggregates all triples that are valid at timestamp $t$. Following the standard bidirectional–relation convention \cite{kazemi2018simple}, we augment every quadruple $(s, r, o, t)$ with its inverse $(o, r^{-1}, s, t)$, where $r^{-1}$ is a distinct relation denoting the reverse semantics of $r$. 

\noindent\textbf{Definition 2. Temporal Knowledge Graph Reasoning. } Let $q = (s, r, ?, t)$ be a query quadruple whose object entity is missing at timestamp $t$. Given the sliding history window of length $L$, $\mathcal{G}_{t-L-1:t-1} = \{\mathcal{G}_{t-L},\mathcal{G}_{t-L+1},\dots,\mathcal{G}_{t-1}\}$. The TKG reasoning task is to learn a scoring function:
\begin{equation}
\footnotesize
\operatorname{score}_t(o)=f(s, r, o, \mathcal{G}),\\
\hat{o}=\underset{o \in \mathcal{E}}{\arg \max } \operatorname{score}_t(o) .
\end{equation}
Each candidate object \(o\in\mathcal{E}\) is assigned a score and the highest‐scoring entity completes the quadruple.

\section{Method}


\begin{figure*}[t]
 \centering
	\includegraphics[scale=0.8]{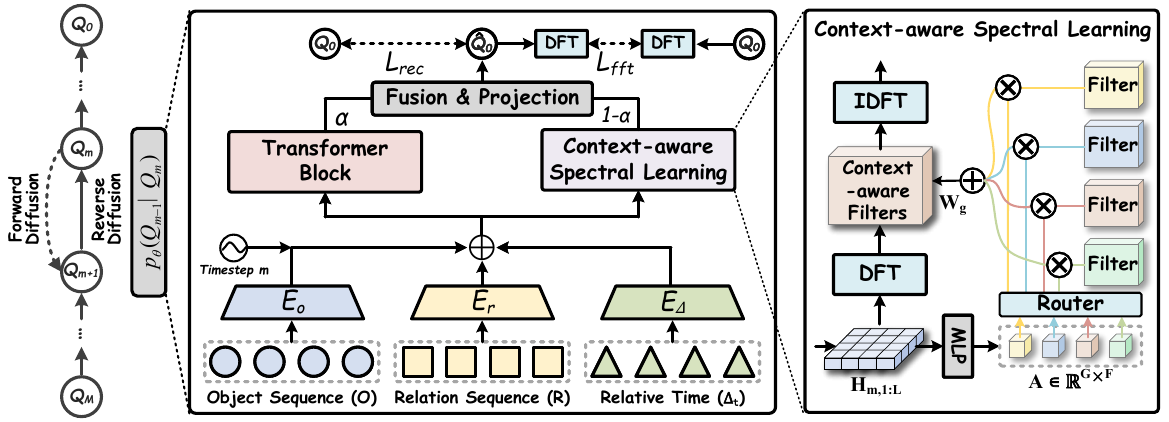}
 \caption{Overview of FreqDiff. Given a subject-oriented history, FreqDiff injects Gaussian noise into the target object representation and reconstructs it through reverse diffusion. The denoiser combines temporal modeling with context-aware spectral calibration, where history-conditioned filters re-calibrate frequency components and fuse them with time-domain representations for future object prediction.}
    \label{fig2*}
\end{figure*}

\subsection{Temporal Representation Learning}

Given a query $q=(s,r_q,?,t)$, the goal is to predict the missing object from the entity set $\mathcal{E}$ based on the recent history of the subject $s$. We first construct a subject-centric sequence comprising the $L$ most recent historical events alongside the query slot:
\begin{equation}
\small
\mathcal{Q}_{s,t}=\big[(s_1,r_1,o_1,\tau_1),\ldots, (s_i,r_i,o_i,\tau_i),(s_L,r_L,o_L,\tau_L)\big].
\end{equation}

Following the standard sequential formulation, we project the object, relation, and relative-time components of $\mathcal{Q}_{s,t}$ into a shared $h$-dimensional space. Let $\mathbf{E}_{o}\in\mathbb{R}^{|\mathcal{E}|\times h}$, $\mathbf{E}_{r}\in\mathbb{R}^{|\mathcal{R}|\times h}$, and $\mathbf{E}_{\Delta t}\in\mathbb{R}^{N_t\times h}$ denote the object, relation, and relative-time embedding matrices, respectively:
\begin{equation}
\small
\begin{aligned}
\mathbf{o} &= [\mathbf{E}_{o}(o_1);\ldots;\mathbf{E}_{o}(o_L);\mathbf{E}_{o}(o_t)],\\
\mathbf{r} &= [\mathbf{E}_{r}(r_1);\ldots;\mathbf{E}_{r}(r_L);\mathbf{E}_{r}(r_q)],\\
\mathbf{t} &= [\mathbf{E}_{\Delta t}(\Delta \tau_1);\ldots;\mathbf{E}_{\Delta t}(\Delta \tau_L);\mathbf{E}_{\Delta t}(1)].
\end{aligned}
\end{equation}
where $\Delta \tau_i$ represents the temporal interval between the $i$-th historical event and the query timestamp $t$.

\subsection{Forward Diffusion Process}
The forward process injects Gaussian noise into the target object representation at the query position. For a sampled diffusion step $m\in\{1,\ldots,M\}$, the corrupted target representation is generated as:
\begin{equation}
\small
\mathbf{o}_m=\sqrt{\bar{\alpha}_m}\,\mathbf{E}_{o}(o_t)+\sqrt{1-\bar{\alpha}_m}\,\boldsymbol{\epsilon},
\quad \boldsymbol{\epsilon}\sim\mathcal{N}(0,\mathbf{I}),
\end{equation}
where $\bar{\alpha}_m=\prod_{j=1}^{m}(1-\beta_j)$ denotes the remaining signal ratio at step $m$. In practice, a linearly scaled accumulated noise schedule is used:
\begin{equation}
\label{eq:linear_schedule}
\small
1 - \bar{\alpha}_m = \delta \cdot \left(\alpha_{\min} + \frac{m - 1}{M - 1} (\alpha_{\max} - \alpha_{\min})\right).
\end{equation}
where $\delta \in [0,1]$ is a global scaling factor that moderates the overall diffusion strength, and $\alpha_{\min}$, $\alpha_{\max}$ bound the noise levels.

\subsection{Dual-stream Denoiser}

During reverse denoising, the model reconstructs the target entity by conditioning on the relational and temporal context of the recent event trajectory, as shown in Figure \ref{fig2*}. To keep the conditioning context deterministic, stochastic corruption is applied only to the query target representation, while the historical object representations remain unchanged. The denoising input is constructed as:
\begin{equation}
\small
\begin{aligned}
\widetilde{\mathbf{O}}_m &= [\mathbf{E}_o(o_1); \ldots; \mathbf{E}_o(o_L); \mathbf{o}_m], \\
\mathbf{X}_m &= \operatorname{LN}\big(\operatorname{Dropout}(\widetilde{\mathbf{O}}_m + \mathbf{R} + \mathbf{T})\big), \\
\mathbf{H}_m &= \operatorname{LN}\big(\mathbf{X}_m + \operatorname{Emb}(m)\big).
\end{aligned}
\end{equation}
where $\widetilde{\mathbf{O}}_m$ contains the observed historical object embeddings at the first $L$ positions and the diffused target representation $\mathbf{o}_m$ at the query position. The relation sequence $\mathbf{R}$, relative-time sequence $\mathbf{T}$, and diffusion-step embedding $\operatorname{Emb}(m)$ are then incorporated to form $\mathbf{H}_m$. The resulting representation is processed by two complementary branches.

\paragraph{Temporal Dependency Learning.}

The temporal branch explicitly models sequential dependencies by processing the conditioned hidden states through a standard Transformer encoder:
\begin{equation}
\small
\mathbf{H}^{\mathrm{time}}_m=\operatorname{Transformer}(\mathbf{H}_m).
\end{equation}

\paragraph{Context-aware Spectral Filtering.}
The spectral branch complements the temporal branch by explicitly modeling frequency-domain variations in the subject-oriented event sequence, which may be less directly preserved by time-domain attention. To generate a context-aware spectral response, the historical representations are summarized by mean pooling and mapped to routing coefficients:
\begin{equation}
\small
\begin{aligned}
\mathbf{c}=\operatorname{Mean-Pool}(\mathbf{H}_{m,1:L})\in\mathbb{R}^{h},\\
\mathbf{A}=\tanh\big(\operatorname{MLP} (\mathbf{c})\big)\in\mathbb{R}^{G\times F},
\end{aligned}
\end{equation}
where $G$ denotes the number of feature groups and $F$ denotes the number of learnable basis filters. Let $\{\mathbf{B}_f\}_{f=1}^F$ be the set of spectral basis filters, where each basis filter satisfies $\mathbf{B}_f\in\mathbb{R}^{K \times (h/G)}$, and $K$ is the number of valid frequency bins produced by FFT. For each feature group $g$, the context-aware spectral filter is synthesized as:
\begin{equation}
\small
\mathbf{W}_g = \sum_{f=1}^{F}
\mathbf{A}_{g,f}\mathbf{B}_f,
\quad
\mathbf{W}_g \in\mathbb{R}^{K \times (h/G)}
\end{equation}
The routing coefficients $\mathbf{A}_{g,f}$ instantiate a context-aware spectral filter by linearly combining the learnable basis filters. The resulting filter $\mathbf{W}_g$ is applied to the group-wise spectrum of $\mathbf{H}_m$ which is split along the feature dimension into $\mathcal{G}$ groups:
\begin{equation}
\small
\begin{aligned}
\mathcal{Z}'_{\mathrm{re},g} = \operatorname{Re}\big(\mathcal{F}(\mathbf{H}_m)_g\big) \odot \mathbf{W}_g,\\
\mathcal{Z}'_{\mathrm{im},g} = \operatorname{Im}\big(\mathcal{F}(\mathbf{H}_m)_g\big) \odot \mathbf{W}_g.
\end{aligned}
\end{equation}
where $\operatorname{Re}(\cdot)$ and $\operatorname{Im}(\cdot)$ denote the real and imaginary spectral components. $g \in \{1,\ldots,G\}$ indexes the feature group, and $\odot$ denotes element-wise multiplication. The calibrated groups are concatenated along the feature dimension and mapped back to the time domain via inverse FFT:
\begin{equation}
\small
\mathbf{H}^{\mathrm{freq}}_m = \mathcal{F}^{-1}(\mathcal{Z}'_{\mathrm{re}} + j \cdot \mathcal{Z}'_{\mathrm{im}}).
\end{equation}
\paragraph{Temporal-spectral Fusion.} The temporal and spectral representations are fused by interpolation:
\begin{equation}
\small
\mathbf{H}^{\mathrm{fuse}}_m = \operatorname{LN} \big(\alpha\mathbf{H}^{\mathrm{time}}_m + (1-\alpha)\mathbf{H}^{\mathrm{freq}}_m \big).
\end{equation}
where $\alpha\in[0,1]$ is a fusion coefficient. The target representation is obtained from the query position:
\begin{equation}
\small
\hat{\mathbf{o}}_0 = \mathbf{H}^{\mathrm{fuse}}_{m,L+1}.
\end{equation}

\subsection{Training Objective}

The model is trained with a reconstruction objective on the denoised query representation, together with an auxiliary frequency-domain regularizer.

\paragraph{Reconstruction Loss.}

Given the denoised target $\hat{\mathbf{o}}_0$, all candidate entities are scored by dot-product matching with the entity embedding matrix $\mathbf{E}_o \in \mathbb{R}^{|\mathcal{E}| \times h}$. The reconstruction loss is defined as the negative log-likelihood of the gold object $o_t$:
\begin{equation}
\small
\mathcal{L}_{\mathrm{rec}} = -\log \left( \operatorname{Softmax}\left( \frac{\mathbf{E}_o \hat{\mathbf{o}}_0}{\sqrt{h}} \right)_{o_t} \right),
\end{equation}

\paragraph{Frequency-Domain Regularization.}
To further constrain the denoising process, a frequency-domain consistency loss is imposed between the denoised target representation and the gold object embedding. Specifically, FFT is applied along the feature dimension of both representations:
\begin{equation}
\small
\begin{aligned}
\mathcal{L}_{\mathrm{fft}} =&\ \ell\Big( \operatorname{Re}\big(\mathcal{F}(\hat{\mathbf{o}}_0)\big),
\operatorname{Re}\big(\mathcal{F}(\mathbf{E}_{o}(o_t))\big)
\Big) + \\ &\ell\Big( \operatorname{Im}\big(\mathcal{F}(\hat{\mathbf{o}}_0)\big),
\operatorname{Im}\big(\mathcal{F}(\mathbf{E}_{o}(o_t))\big)
\Big),
\end{aligned}
\end{equation}
where $\ell(\cdot,\cdot)$ denotes a point-wise distance function\footnote{Here we utilize point-wise L1 distance.}. This regularizer encourages the denoised representation to align with the gold object not only in the embedding space, but also in its feature-domain spectral profile, providing an explicit constraint for frequency-aware denoising.

The overall training objective is then denoted as:
\begin{equation}
\small
\mathcal{L} = \mathcal{L}_{\mathrm{rec}} + \lambda\mathcal{L}_{\mathrm{fft}}.
\end{equation}
where $\lambda$ controls the strength of the auxiliary frequency-domain regularization.

\subsection{Inference}
During inference, FreqDiff initialize the unknown target representation with Gaussian noise. A standard reverse diffusion sampler would apply the denoising network $f_\theta(\cdot)$ step by step from $M$ to 0, but this iterative process increases computational cost. Since $f_\theta(\cdot)$ is trained to recover the clean target representation from a corrupted representation $\mathrm{o}_m$ at any diffusion step $m$, we adopt an efficient inference strategy \cite{cai-etal-2024-predicting,gan2026negative} that directly predicts the clean target from the maximum-noise representation $\mathrm{o}_M$, without performing all intermediate reverse transitions:
\begin{equation}
\small
\begin{gathered}
\widetilde{\mathbf{O}}_M=[\mathbf{E}_o(o_1); \ldots; \mathbf{E}_o(o_L); \mathbf{o}_M],\\
\mathbf{H}_M=\widetilde{\mathbf{O}}_M + \mathbf{R} + \mathbf{T}+\operatorname{Emb}(M).
\end{gathered}
\end{equation}
The learned denoising network then directly predicts the clean target representation:
\begin{equation}
\small
\hat{\mathbf{o}}_0 = f_{\theta}(\mathbf{H}_M, M).
\end{equation}

\section{Experiments}

\subsection{Experimental Setups}

\noindent\textbf{Datasets.} Our experiments employ four benchmark datasets, including ICEWS14, ICEWS05-15, ICEWS18, and GDELT, to evaluate the proposed model. Specifically, the ICEWS datasets originate from the Integrated Crisis Early Warning System \cite{Boschee15ICEWS}, while the GDELT dataset is sourced from the Global Database of Events, Language, and Tone \cite{Leetaru13gdelt}. The data statistics are summarized in Appendix \ref{data stat}.

\noindent\textbf{Baseline Models.} We benchmark FreqDiff against three sets of approaches: \textbf{Static methods}: DistMult \cite{yang2015embedding}, ConvE \cite{dettmers2018convolutional}, RotatE\cite{sun2018rotate}; \textbf{Interpolation methods}: TTransE \cite{leblay2018deriving}, TA-DistMult \cite{garcia2018learning}, DE-SimpIE \cite{goel2020diachronic}; \textbf{Extrapolation methods}: RE-NET \cite{jin2020recurrent}, Re-GCN \cite{Li21Temporal}, CEN \cite{li2022complex}, TiRGN \cite{li2022tirgn}, TITer \cite{sun2021timetraveler}, RETIA \cite{liu2023retia}, CENET \cite{xu2023temporal}, THCN \cite{chen2024thcn}, DiffuTKG \cite{cai-etal-2024-predicting}, LogiQ \cite{chen2025enhancing}, CognTKE \cite{chen2025cogntke}, NADEx \cite{gan2026negative}. We provide detailed baseline descriptions in Appendix \ref{baseline}. 


\noindent\textbf{Evaluation Metrics.} To measure temporal extrapolation performance, we cast the task as masked entity prediction, where either the subject or object is held out in quadruples of the form $(s,r,?,t)$ or $(?,r,o,t)$. Predictions are scored and ranked, and we report Mean Reciprocal Rank (MRR) alongside Hits@1, Hits@3, and Hits@10. All results are computed under the time-aware filtering protocol.

\noindent\textbf{Implementation Details.} All models are optimized with Adam and trained for 100 epochs. The learning rate is set to $1e^{-3}$ on ICEWS14 and ICEWS18, and $5e^{-4}$ on ICEWS05-15 and GDELT. Entity and relation embeddings are both initialized with a dimensionality of 200. Experiments are conducted on a single NVIDIA A100 GPU with 80GB memory. Hyper-parameter configurations are provided in Appendix \ref{implemt}. \textit{The reported results are averaged with five runs with different seeds.}

\begin{table*}[t]
\caption{Performance comparison (\%) on four benchmarks with MRR and Hits@1/3/10. Best and second-best results are shown in \textbf{bold} and \underline{underlined}, respectively. $\spadesuit$ indicates results re-implemented using official code.}
\setlength{\tabcolsep}{2pt} 
\fontsize{7.5pt}{6pt}\selectfont 
\begin{tabular}{@{}ccccccccccccccccc@{}}
\toprule
\multirow{2}{*}{Models} & \multicolumn{4}{c}{ICEWS14}    & \multicolumn{4}{c}{ICEWS18}    & \multicolumn{4}{c}{ICEWS05-15} & \multicolumn{4}{c}{GDELT} \\ \cmidrule(l){2-17} 
  & MRR   & Hit@1 & Hit@3 & Hit@10 & MRR   & Hit@1 & Hit@3 & Hit@10 & MRR   & Hit@1 & Hit@3 & Hit@10 & MRR   & Hit@1 & Hit@3 & Hit@10 \\ \midrule
DisMult \citeyearpar{yang2015embedding}  & 15.44 & 10.91 & 17.24 & 23.92  & 11.51 & 7.03  & 12.87 & 20.86  & 17.95 & 13.12 & 20.71 & 29.32  & 8.68  & 5.58  & 9.96  & 17.13  \\
ConvE \citeyearpar{dettmers2018convolutional}    & 35.09 & 25.23 & 39.38 & 54.68  & 24.51 & 16.23 & 29.25 & 44.51  & 33.81 & 24.78 & 39.00 & 54.95  & 16.55 & 11.02 & 18.88 & 31.60  \\
RotatE \citeyearpar{sun2018rotate}    & 21.31 & 10.26 & 24.35 & 44.75  & 12.78 & 4.01  & 14.89 & 31.91  & 24.71 & 13.22 & 29.04 & 48.16  & 13.45 & 6.95  & 14.09 & 25.99  \\ \midrule
TTransE \citeyearpar{leblay2018deriving}  & 13.72 & 2.98  & 17.70 & 35.74  & 8.31  & 1.92  & 8.56  & 21.89  & 15.57 & 4.80  & 19.24 & 38.29  & 5.50  & 0.47  & 4.94  & 15.25  \\
TA-DisMult \citeyearpar{garcia2018learning}  & 25.80 & 16.94 & 29.74 & 42.99  & 16.75 & 8.61  & 18.41 & 33.59  & 24.31 & 14.58 & 27.92 & 44.21  & 12.00 & 5.76  & 12.94 & 23.54  \\
DE-SimIE \citeyearpar{goel2020diachronic}  & 33.36 & 24.85 & 37.15 & 48.92  & 19.30 & 11.53 & 21.86 & 34.80  & 35.02 & 25.91 & 38.99 & 52.75  & 19.70 & 12.22 & 21.39 & 33.70  \\ \midrule
RE-NET \citeyearpar{jin2020recurrent}    & 36.93 & 26.83 & 39.51 & 54.78  & 28.81 & 19.05 & 32.44 & 47.51  & 43.32 & 33.43 & 47.77 & 63.06  & 19.62 & 12.42 & 21.00 & 34.01  \\
RE-GCN \citeyearpar{Li21Temporal}    & 40.39 & 30.66 & 44.96 & 59.21  & 30.58 & 21.01 & 34.34 & 48.75  & 48.03 & 37.33 & 53.85 & 68.27  & 19.64 & 12.42 & 20.90 & 33.69  \\
CyGNet \citeyearpar{zhu2021learning}    & 35.05 & 25.73 & 39.01 & 53.55  & 24.93 & 15.90 & 28.28 & 42.61  & 36.81 & 26.61 & 41.63 & 56.22  & 18.48 & 11.52 & 19.57 & 31.98  \\
TITer \citeyearpar{sun2021timetraveler}    & 41.73 & 32.74 & 46.46 & 58.44  & 29.98 & 22.05 & 33.46 & 44.83  & 47.69 & 37.95 & 52.92 & 65.81  & 15.46 & 10.98 & 15.61 & 24.31  \\
CEN \citeyearpar{li2022complex}  & 42.20 & 32.08 & 47.46 & 61.31  & 31.50 & 21.70 & 35.44 & 50.59  & 46.84 & 36.38 & 52.45 & 67.01  & 20.39 & 12.96 & 21.77 & 34.97  \\
TiRGN \citeyearpar{li2022tirgn} & 44.04 & 33.83 & 48.95 & 63.84  & 33.66 & 23.19 & 37.99 & 54.22  & 50.04 & 39.25 & 56.13 & 70.71  & 21.67 & 13.63 & 23.27 & 37.60  \\
RETIA \citeyearpar{liu2023retia} & 42.76 & 32.28 & 47.77 & 62.75  & 32.43 & 22.23 & 36.48 & 52.94  & 47.26 & 36.64 & 52.90 & 67.76  & 20.12 & 12.76 & 21.45 & 34.49  \\
CENET \citeyearpar{xu2023temporal} & 39.02 & 29.62 & 43.23 & 57.49  & 27.85 & 18.15 & 31.63 & 46.98  & 41.95 & 32.17 & 46.93 & 60.43  & 20.23 & 12.69 & 21.70 & 34.92  \\
THCN \citeyearpar{chen2024thcn} & 45.39 & 36.58 & 50.84 & 66.07  & 35.63 & 24.90 & 39.26 & 56.76  & 51.94 & 40.32 & 57.79 & 72.18  & \underline{23.46} & \underline{15.18} & \underline{25.21} & \underline{39.03}  \\
DiffuTKG$^\spadesuit$ \citeyearpar{cai-etal-2024-predicting} & 47.58 &   36.38    &    53.41   & 66.01   & \underline{35.65} & 25.19  &   39.39    & \underline{59.55}   & 48.97  &   39.80    &  56.92 &    69.84    &   21.35    &  14.43 &   23.68    &  36.05 \\
LogiQ \citeyearpar{chen2025enhancing} & 44.71 & 35.72 & 51.03 & 64.21  & 34.94 & 24.76 & 39.57 & 56.32  & 51.04 & 40.71 & 57.55 & 71.00  & -- & -- & -- & --  \\
CognTKE \citeyearpar{chen2025cogntke} & 46.06 & 36.49 & 51.11 & 64.49  & 35.24 & 25.21 & 39.93 & 54.71  & \underline{53.13} & 42.62 & 59.42 & \underline{72.70}  & -- & -- & -- & --  \\
NADEx$^\spadesuit$ \citeyearpar{gan2026negative} & \underline{48.12}  & \underline{37.89} & \underline{55.26} & \underline{70.55}  & 35.37  & \underline{25.48}  & \underline{40.27} & 58.69  & 52.17  & \underline{43.38} & \underline{60.47} & 71.93  & 21.78 & 14.69 & 23.37 & 37.17  \\ \midrule
FreqDiff & \textbf{50.85}  & \textbf{39.82} & \textbf{57.48} & \textbf{71.48}  & \textbf{36.85}  & \textbf{26.85}  & \textbf{41.51} & \textbf{61.47}  & \textbf{54.71}  & \textbf{45.02} & \textbf{61.16} & \textbf{74.45}  & \textbf{25.04} & \textbf{17.64} & \textbf{27.03} & \textbf{41.83}  \\
\textit{Improve.} & 5.67\%  & 5.09\% & 4.01\% & 1.32\%  & 3.37\%  & 5.37\%  & 3.08\% & 3.22\%  & 2.97\%  & 3.78\% & 1.14\% & 2.40\%  & 6.73\% & 16.20\% & 7.22\% & 7.17\%  \\ \bottomrule
\end{tabular}\label{main}
\end{table*}

\subsection{Overall Performance}
Table \ref{main} summarizes FreqDiff's performance against state‑of‑the‑art (SOTA) baselines across four benchmark datasets. From these results, we make the following key observations:
\begin{itemize}[leftmargin=*, itemsep=0pt, parsep=0pt]
    \item \textbf{FreqDiff achieves the best results across all four datasets and all 16 evaluation metrics}, with relative improvements ranging from 1.14\% to 16.14\% over the strongest baseline. This consistent superiority demonstrates that incorporating frequency-aware modeling into the diffusion framework leads to broadly effective temporal knowledge graph extrapolation.
    
    \item \textbf{The largest improvement appears on GDELT}, especially with a 16.14\% gain on Hit@1 and a 6.73\% gain on MRR. Since GDELT involves more complex temporal evolution patterns, these gains suggest that frequency-aware modeling is beneficial in challenging extrapolation scenarios. This further supports that spectral information provides complementary signals beyond conventional temporal modeling.

    \item \textbf{Static and interpolation-based methods lag behind extrapolation-oriented models}, confirming that forward-looking temporal reasoning is essential for predicting future facts. Although interpolation models can exploit temporal information within observed histories, they are not explicitly optimized for future event evolution, which limits their ability in extrapolation settings.

\end{itemize}

\begin{table}[t]
\setlength{\tabcolsep}{4.5pt} 
\fontsize{7.5pt}{6pt}\selectfont 
\caption{Ablation study results ICEWS14 and ICEWS18 datasets in terms of MRR and Hit@1/10.}
\begin{tabular}{@{}ccccccc@{}}
\toprule
\multirow{2}{*}{Settings} & \multicolumn{3}{c}{ICEWS14} & \multicolumn{3}{c}{ICEWS18} \\ \cmidrule(l){2-7} 
 & MRR   & Hit@1   & Hit@10   & MRR   & Hit@1   & Hit@10  \\ \midrule
FreqDiff  & \textbf{50.85} & \textbf{39.82} & \textbf{71.48} & \textbf{36.85} & \textbf{26.85} & \textbf{61.47} \\ \midrule
w/o. $Freq$  & 48.68 & 38.94 & 70.64 & 35.81 & 26.04 & 58.79 \\
w/o. $Time$  & 42.14 & 32.37 & 65.39 &   18.21    &    11.76 &   35.52 \\
w/o. $\mathcal{L}_\text{fft}$  & 49.30 & 38.84 & 69.82 &    36.10   &  26.13  & 59.47    \\ 
w. $\ell_{2}$ Distance   & 50.21 & 39.13 & 70.77 & 35.14 & 25.53 & 59.22 \\\midrule
w. global filter   & 48.70 & 39.03 & 70.58 & 34.42 & 24.93 & 58.01 \\
w. static filter   & 48.61 & 38.89 & 69.81 & 32.07 & 23.10 & 55.18 \\
w. maxpool   & 49.06 & 39.46 & 70.01 & 34.05 & 24.84 & 57.78 \\
 \bottomrule
\end{tabular}\label{ablation}
\end{table}

\subsection{Ablation Studies}
We validate the contribution of each FreqDiff component by comparing it against seven variants:

\begin{itemize}[leftmargin=*, itemsep=0pt, parsep=0pt]
    \item \textbf{w/o. Freq:} omits frequency-domian modeling.
    \item \textbf{w/o. Time:} no time-domain modeling.
    \item \textbf{w/o. $\mathcal{L}_\text{fft}$:} removes frequency-domain loss.
    \item \textbf{w. $\ell_{2}$ Distance}: replaces $\ell_{1}$ distance in the frequency-domain regularizer with $\ell_{2}$ distance.
    \item \textbf{w. global filter:} replaces context-aware spectral filter with a shared global filter.
    \item \textbf{w. static filter:} replaces context-aware spectral filter with a static filter.
    \item \textbf{w. maxpool:} replaces mean-pool with maxpool.
\end{itemize}

As shown in Table \ref{ablation}, both time-domain and frequency-domain designs are necessary for FreqDiff. Removing the frequency branch (w/o. $Freq$) causes consistent drops, confirming that spectral information provides useful complementary signals. Removing the time branch (w/o. $Time$) leads to the most severe degradation, indicating that time-domain modeling remains the core component for preserving event order and relation-specific temporal context. In contrast, frequency modeling works as a complementary signal. The drop after removing $\mathcal{L}_\text{fft}$ further verifies the value of explicit spectral supervision. Moreover, replacing the $\ell{1}$ distance with $\ell_{2}$ leads to inferior results, suggesting that $\ell_{1}$ provides a more robust constraint for spectral alignment. Finally, replacing the context-aware filter with global or static filters consistently weakens performance, suggesting that different event histories require adaptive frequency responses rather than shared or fixed filtering. The degradation caused by max-pooling further shows that, within the frequency branch, mean-pooling better preserves the overall spectral distribution, whereas max-pooling may overemphasize dominant frequency components and suppress weaker but informative spectral patterns.

\begin{figure}[t]
	\centering
	\subfloat[FreqDiff w/o. Freq]{
		\includegraphics[scale=0.45]{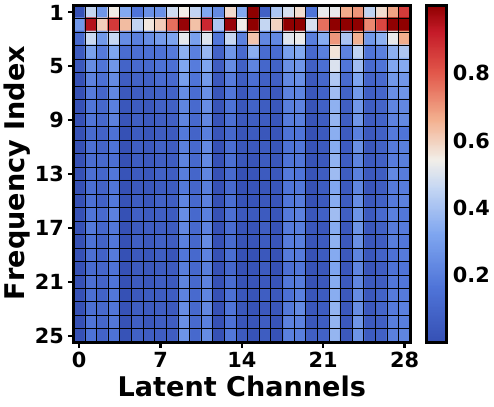}
	}%
 	\subfloat[FreqDiff]{
		\includegraphics[scale=0.45]{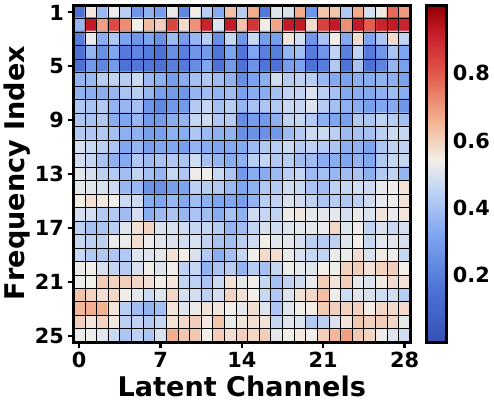}
	}%
	\centering

    \caption{Visualization of learned spectral energy distributions on ICEWS14. The y-axis represents frequency indices, and the x-axis represents latent channels. \textcolor{red}{Red} indicates higher spectral energy.}
	\label{Spec-viz}
\end{figure}

\begin{figure}[t]
	\centering
	\subfloat[Impact on ICEWS18]{
		\includegraphics[scale=0.185]{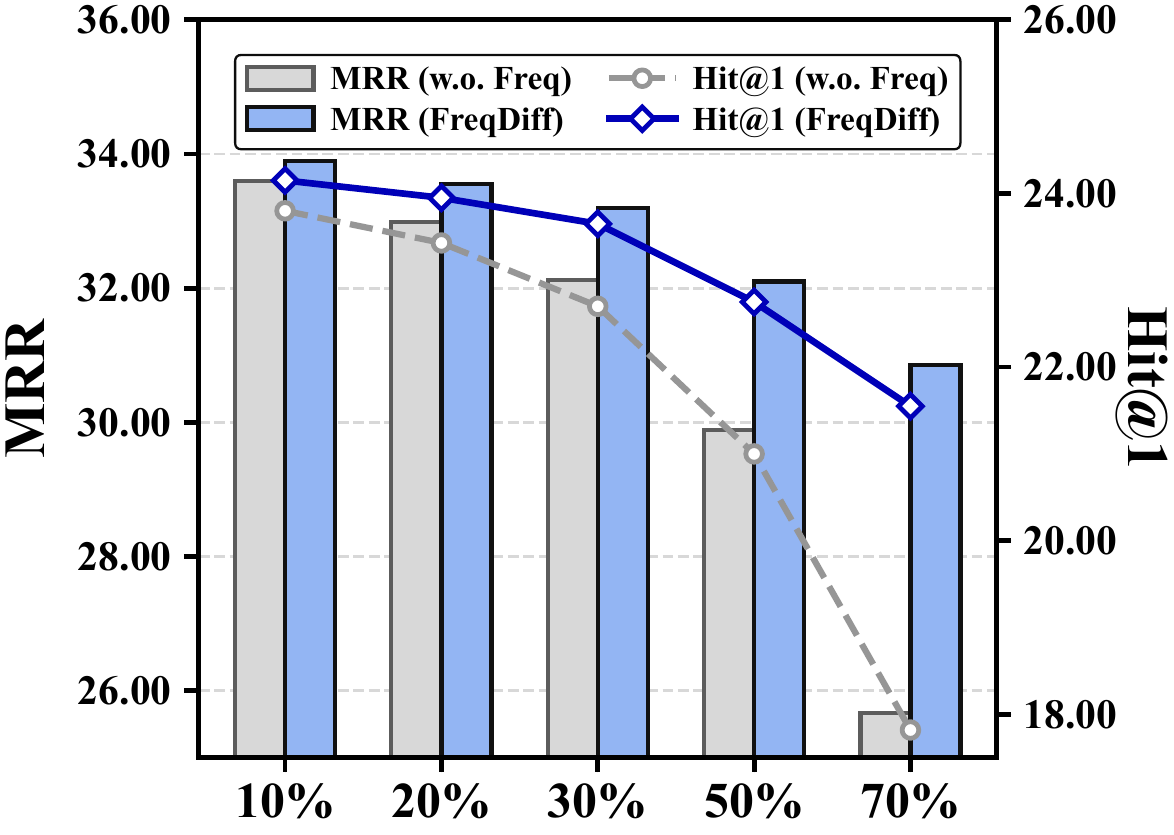}
	}%
 	\subfloat[Impact on GDELT]{
		\includegraphics[scale=0.185]{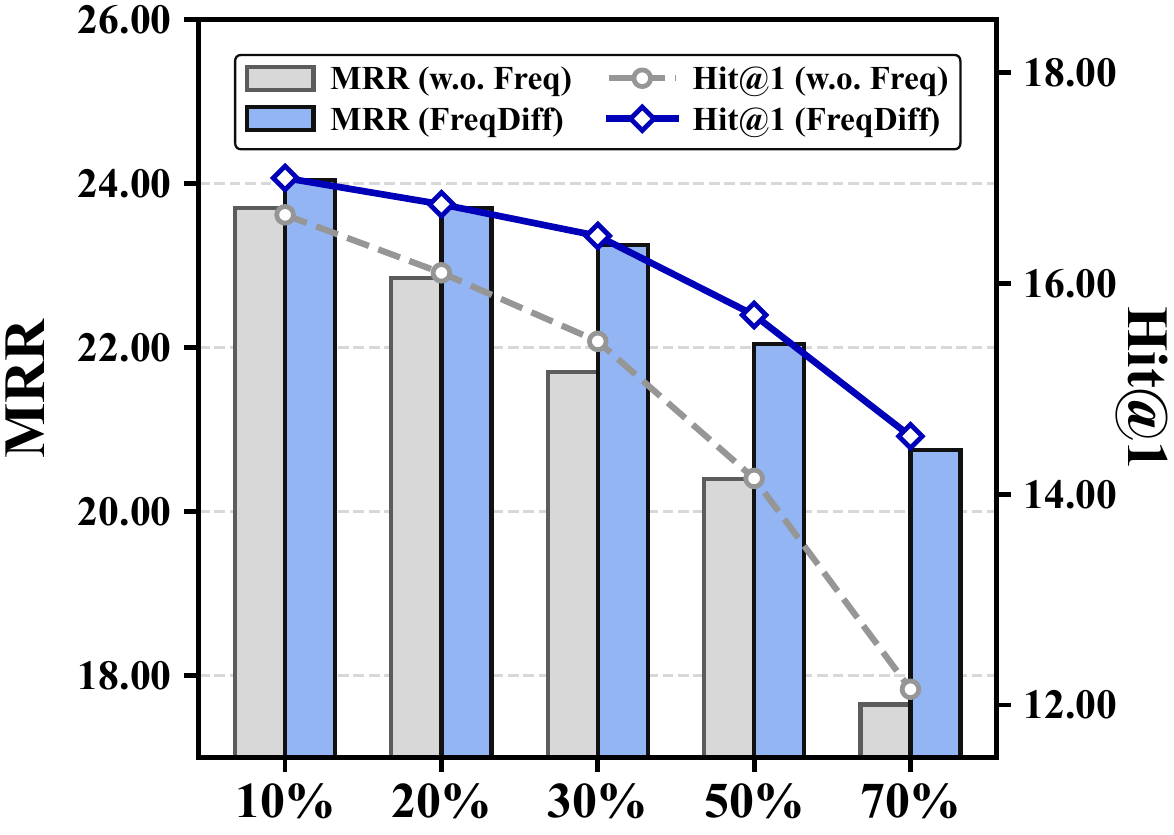}
	}%
	\centering

    \caption{Performance under different ratios of sequence corruption on ICEWS18 and GDELT.}
	\label{Noise}
\end{figure}

\subsection{Analysis of Spectral Learning}
We further examine how spectral learning contributes to contextual representation learning and performance under sequence corruption.

\paragraph{Visualization of Spectral Learning.} Figure \ref{Spec-viz} visualizes the learned spectral energy distributions of FreqDiff and its variant without frequency modeling. The variant without frequency modeling shows a relatively concentrated response pattern, suggesting that its denoising representation is mainly dominated by the overall temporal patterns in the subject history. In contrast, FreqDiff exhibits a more adaptive spectral response across frequency indices and latent channels. Rather than uniformly amplifying spectral responses, it modulates the frequency-channel components of the denoising representation, attenuating generic temporal signals while retaining components more aligned with the queried relation and the target object. In this way, spectral learning introduces finer query-conditioned temporal variations into the denoising process. Additional visualized comparisons with other diffusion-based models are provided in Appendix \ref{D.1}.

\paragraph{Effectiveness under Noise Condition.} Figure \ref{Noise} further evaluates the effectiveness of FreqDiff when the historical context is partially corrupted. Specifically, for each test query, we randomly select a given proportion (ranging from 10\% to 70\%) of historical events in the subject-centric sequence and perturb their event representations while keeping the trained model unchanged. As the corruption ratio increases, the performance of both FreqDiff and its variant without spectral learning declines, indicating that reliable historical context is important for TKG extrapolation. Nevertheless, FreqDiff consistently outperforms the variant without spectral modeling on both ICEWS18 and GDELT in terms of MRR and Hits@1, with a more evident advantage under higher corruption ratios. These results suggest that spectral learning improves robustness under corrupted historical observations. By introducing a frequency-aware filter, FreqDiff re-calibrates temporal-frequency representations, attenuating corrupted responses while preserving reliable denoising signals.

\begin{table}[t]
\caption{Effectiveness of the frequency regularizer. \textcolor{red}{Red} superscripts indicate the improvement rates.}
\setlength{\tabcolsep}{1pt}
\fontsize{7pt}{8pt}\selectfont
\resizebox{\linewidth}{!}{%
\begin{tabular}{@{}cccccccc@{}}
\toprule
\multirow{2}{*}{Dataset} & \multirow{2}{*}{Metric} 
& \multicolumn{2}{c}{DiffuTKG} 
& \multicolumn{2}{c}{NADEx} 
& \multicolumn{2}{c}{CENET} \\ 
\cmidrule(lr){3-4} \cmidrule(lr){5-6} \cmidrule(l){7-8}
& & Base & + $\mathcal{L}_{\mathrm{fft}}$ 
  & Base & + $\mathcal{L}_{\mathrm{fft}}$ 
  & Base & + $\mathcal{L}_{\mathrm{fft}}$ \\ 
\midrule

\multirow{2}{*}{ICEWS14} 
& MRR   
& 47.58 
& 48.67\textsuperscript{\textcolor{red}{+2.29\%}} 
& 48.12 
& 49.03\textsuperscript{\textcolor{red}{+1.89\%}} 
& 39.02 
& 43.39\textsuperscript{\textcolor{red}{+11.20\%}} \\

& H@1 
& 36.38 
& 37.73\textsuperscript{\textcolor{red}{+3.71\%}} 
& 37.89 
& 38.97\textsuperscript{\textcolor{red}{+2.85\%}} 
& 29.62 
& 32.08\textsuperscript{\textcolor{red}{+8.31\%}} \\ 

\midrule

\multirow{2}{*}{ICEWS18} 
& MRR   
& 35.65 
& 36.24\textsuperscript{\textcolor{red}{+1.65\%}} 
& 35.37 
& 36.04\textsuperscript{\textcolor{red}{+1.89\%}} 
& 27.85 
& 29.96\textsuperscript{\textcolor{red}{+7.58\%}} \\

& H@1 
& 25.19 
& 26.36\textsuperscript{\textcolor{red}{+4.64\%}} 
& 25.48 
& 26.13\textsuperscript{\textcolor{red}{+2.55\%}} 
& 18.15 
& 20.85\textsuperscript{\textcolor{red}{+14.88\%}} \\ 

\midrule

\multirow{2}{*}{GDELT} 
& MRR   
& 21.35 
& 22.90\textsuperscript{\textcolor{red}{+7.26\%}} 
& 21.78 
& 23.18\textsuperscript{\textcolor{red}{+6.43\%}} 
& 20.23 
& 21.90\textsuperscript{\textcolor{red}{+8.26\%}} \\

& H@1 
& 14.43 
& 15.72\textsuperscript{\textcolor{red}{+8.94\%}} 
& 14.69 
& 15.45\textsuperscript{\textcolor{red}{+5.17\%}} 
& 12.69 
& 14.01\textsuperscript{\textcolor{red}{+10.40\%}} \\ 

\bottomrule
\end{tabular}%
}
\label{loss}
\end{table}

\subsection{Generalization Analysis}

\paragraph{Effectiveness of Frequency Regularizer.}
Table \ref{loss} validates the effectiveness and \textit{plug-and-play} applicability of the frequency-domain regularizer across different backbone models and datasets. Without changing the original architectures or training settings, incorporating $\mathcal{L}_\text{fft}$ consistently improves DiffuTKG, NADEx, and CENET on ICEWS14, ICEWS18, and GDELT under both MRR and H@1. Specifically, $\mathcal{L}_\text{fft}$ improves DiffuTKG, NADEx, and CENET by 3.73\%/5.76\%, 3.40\%/3.52\%, and 9.01\%/11.20\% on average in terms of MRR/H@1, respectively. These improvements confirm that penalizing frequency-domain inconsistency provides complementary guidance to conventional objectives, leading to more discriminative representations for event prediction.

\begin{table}[t]
\caption{Performance of predicting unseen events in terms of MRR and Hit@1 on ICEWS14 and ICEWS18.}
\setlength{\tabcolsep}{12pt} 
\fontsize{8pt}{6pt}\selectfont 
\begin{tabular}{@{}ccccc@{}}
\toprule
\multirow{2}{*}{Models} & \multicolumn{2}{c}{ICEWS14} & \multicolumn{2}{c}{ICEWS18} \\ \cmidrule(l){2-5} 
  & MRR  & Hit@1 & MRR  & Hit@1 \\ \midrule
RE-GCN    & 23.26 & 13.91 & 15.08 & 7.09 \\
CEN  & 22.06 & 13.28 & 15.41 & 8.20 \\
RETIA & 24.17 & 14.67 & 16.62 & 9.08 \\
HisMatch  & 27.49 & 19.04 & 17.51 & 11.13 \\
DiffuTKG  & 25.22 & 15.23 & 16.48 & 8.84 \\
NADEx & \underline{29.71} & \underline{19.34} & \underline{19.52} & \underline{12.17} \\ \midrule
FreqDiff & \textbf{31.42} & \textbf{21.33} & \textbf{20.33} & \textbf{12.55} \\
\textit{Improve.} & \textit{5.75\%} & \textit{10.29\%} & \textit{4.15\%} & \textit{3.12\%} \\ \bottomrule
\end{tabular}\label{unseen}
\end{table}

\paragraph{Effectiveness on Unseen Events.} To examine FreqDiff's generalization to unseen temporal relational facts, we evaluate unseen event prediction on ICEWS14 and ICEWS18. As shown in Table \ref{unseen}, FreqDiff achieves the best results across all metrics, outperforming the strongest baseline NADEx by 5.75\%/10.29\% on ICEWS14 and 4.15\%/3.12\% on ICEWS18 in terms of MRR/Hit@1. The consistent gains over conventional temporal reasoning methods and diffusion-based baselines suggest that frequency-aware modeling offers complementary signals for unseen event prediction. By exploiting spectral patterns beyond time-domain histories, FreqDiff better distinguishes plausible future facts from unseen candidates.

\begin{figure}[t]
	\centering
    \hspace*{-0.3cm}%
	\subfloat[Impact of $G$]{
		\begin{tabular}{@{}c@{\hspace{0.2cm}}c@{}}
			\includegraphics[scale=0.16]{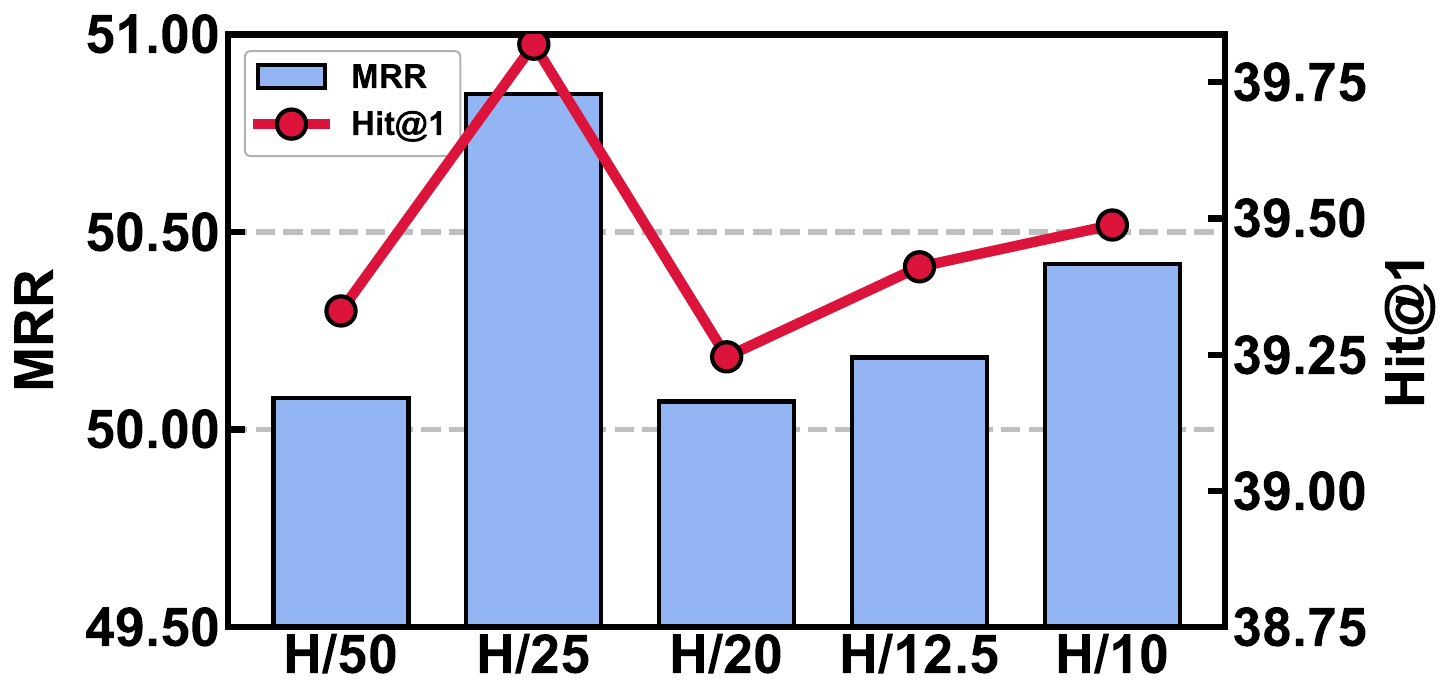} &
			\includegraphics[scale=0.16]{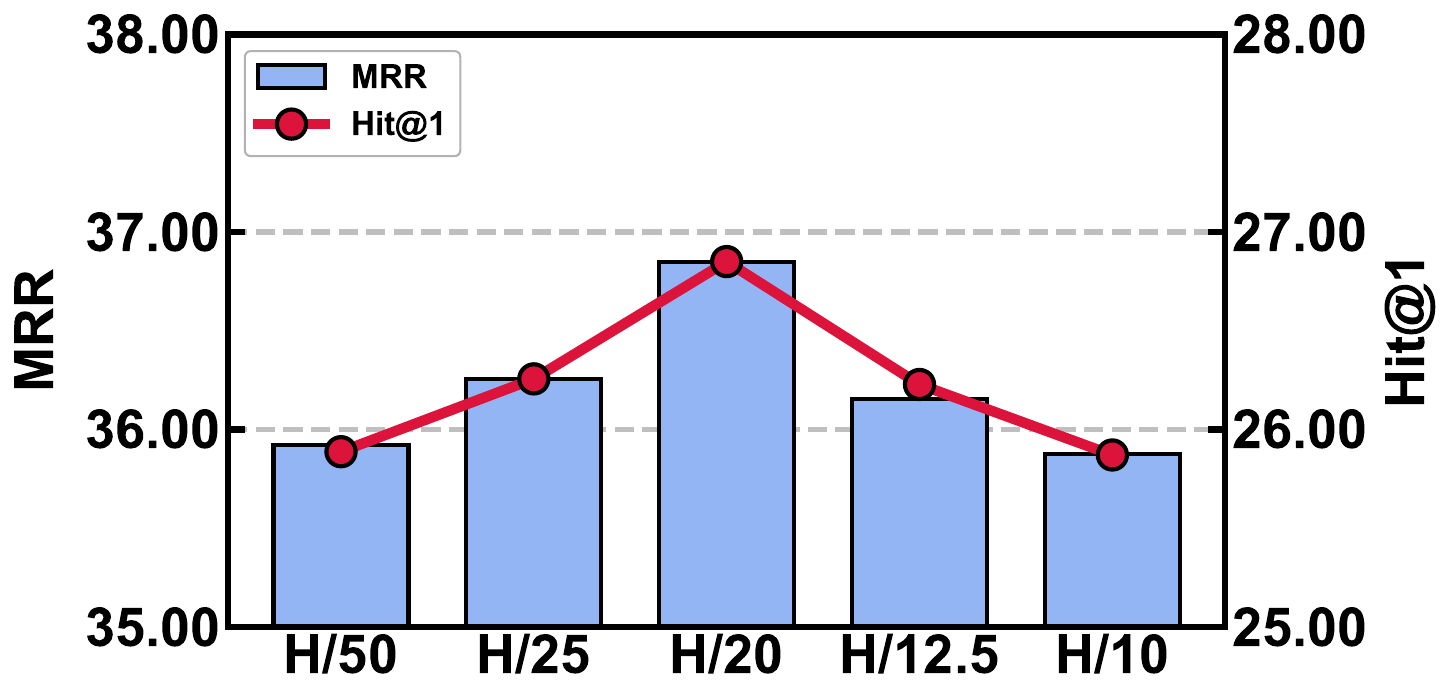}
		\end{tabular}
	}\\

    \hspace*{-0.3cm}%
	\subfloat[Impact of $F$]{
		\begin{tabular}{@{}c@{\hspace{0.2cm}}c@{}}
			\includegraphics[scale=0.16]{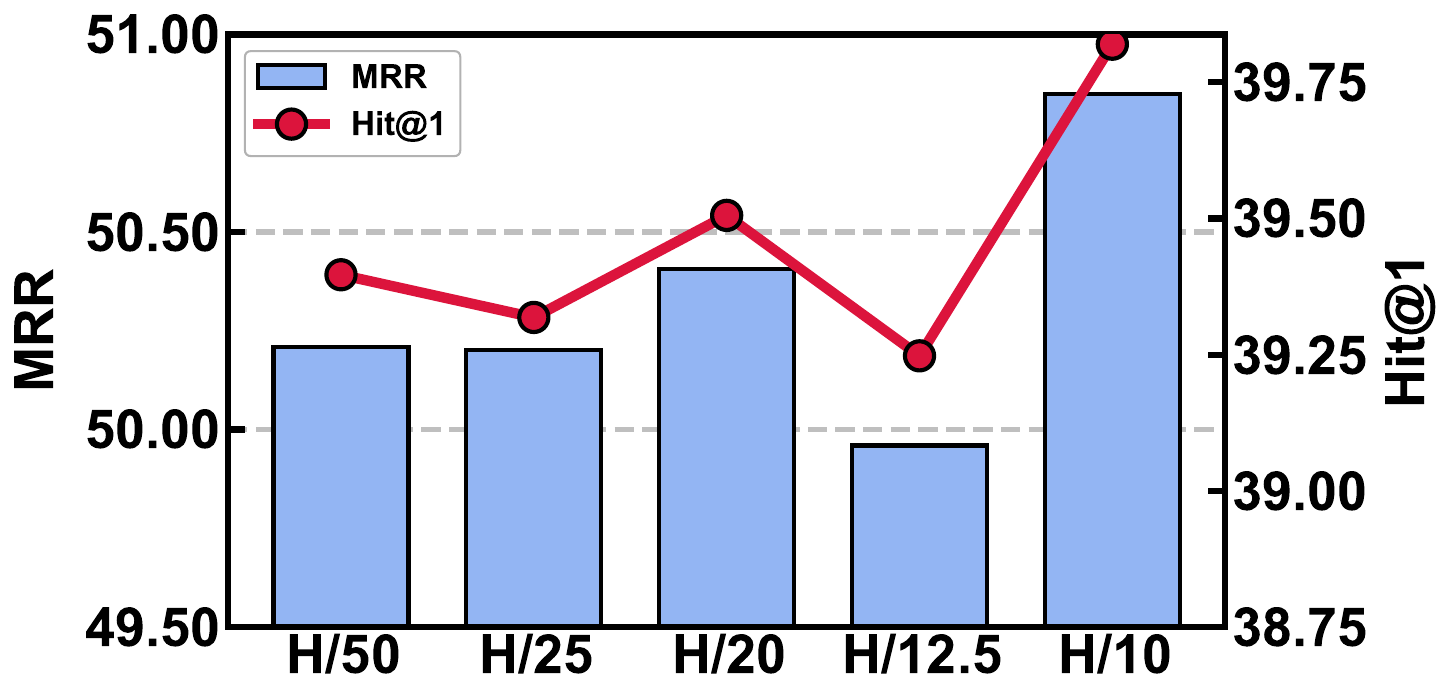} &
			\includegraphics[scale=0.16]{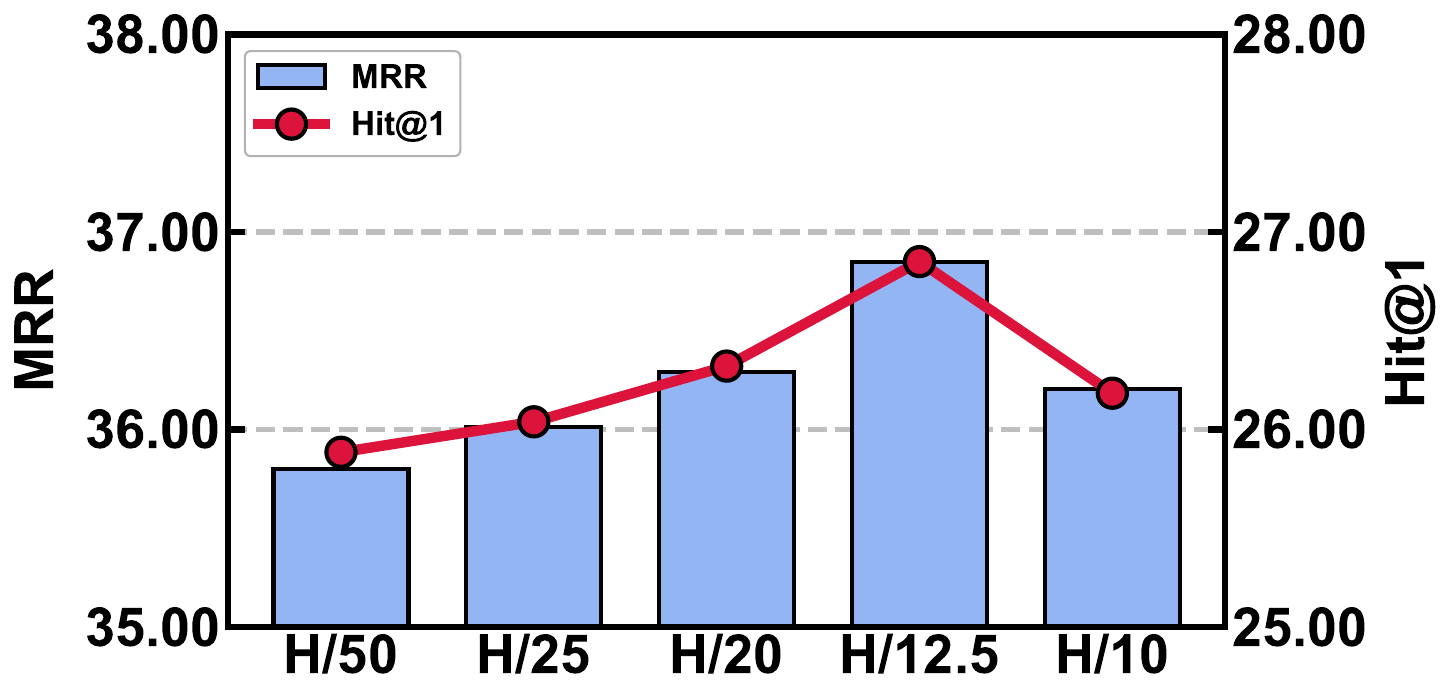}
		\end{tabular}
	}

	\caption{Hyper-parameter sensitivity analysis on ICEWS14 (left) and ICEWS18 (right) datasets.}
	\label{Hyper}
\end{figure}

\subsection{Sensitivity Analysis}
Figure \ref{Hyper} shows that FreqDiff is generally stable across different hyper-parameter settings. For the feature groups $G$, the best results appear around H/25 on ICEWS14 and H/20 on ICEWS18, suggesting that moderate grouping provides a better balance between flexible spectral calibration and stable representation learning. For the basis number $F$, performance remains robust, while larger or moderate basis sets usually work better, indicating that multiple spectral bases help capture diverse frequency patterns. Overall, the results indicate that FreqDiff is not overly sensitive to hyper-parameters, and its best performance comes from a balanced temporal-spectral configuration. Additional analysis on fusion coefficient $\alpha$ and loss balance term $\lambda$ is provided in Appendix \ref{D.3}.

\begin{table}[]
\caption{Computational efficiency test on ICEWS14/18.}
\setlength{\tabcolsep}{8pt} 
\fontsize{8pt}{10pt}\selectfont 
\begin{tabular}{@{}ccccc@{}}
\toprule
\multirow{2}{*}{Model} & \multicolumn{2}{c}{ICEWS14} & \multicolumn{2}{c}{ICEWS18} \\ \cmidrule(l){2-5} 
 & Inf. Time & Params. & Inf. Time & Params. \\ \midrule
RE-GCN   & 28.96s & 26.52Mb & 393.67s & 42.68Mb \\
TiRGN    & 32.88s & 43.35Mb & 159.90s & 59.59Mb \\
NADEx    & 10.91s  & 16.30Mb    & 96.95s & 32.42Mb    \\ \midrule
\textbf{FreqDiff}    & \textbf{7.28s}  & 17.00Mb    & \textbf{81.24s} & 32.83Mb    \\
\bottomrule
\end{tabular}\label{comp}
\end{table}

\subsection{Computational Efficiency}
Table \ref{comp} compares the inference efficiency and parameter scale of different models on ICEWS14 and ICEWS18 on a single NVIDIA A100 GPU under their optimal settings. FreqDiff achieves the fastest inference on both datasets, reducing inference time from 10.91s to 7.28s on ICEWS14 and from 96.95s to 81.24s on ICEWS18 compared with NADEx, corresponding to reductions of 33.27\% and 16.20\%, respectively. Meanwhile, FreqDiff only introduces a slight parameter increase over NADEx, from 16.30Mb to 17.00Mb on ICEWS14 and from 32.42Mb to 32.83Mb on ICEWS18. These results show that the frequency-aware design improves predictive performance while maintaining competitive efficiency, achieving a favorable balance among accuracy, inference speed, and model size.

\subsection{Comparison with LLM-based Forecasters}

\begin{table*}[t]
\centering
\caption{Performance comparison (\%) with LLM-based forecasters. The best results are highlighted in \textbf{bold}.}
\label{tab:tkg-results}
\setlength{\tabcolsep}{3.2pt}
\renewcommand{\arraystretch}{1}
\resizebox{\textwidth}{!}{%
\begin{tabular}{@{}ccccccccccccccccc@{}}
\toprule
\multirow{2}{*}{\textbf{Model}}
& \multicolumn{4}{c}{\textbf{ICEWS14}}
& \multicolumn{4}{c}{\textbf{ICEWS05-15}}
& \multicolumn{4}{c}{\textbf{ICEWS18}}
& \multicolumn{4}{c}{\textbf{GDELT}} \\
\cmidrule(lr){2-5}
\cmidrule(lr){6-9}
\cmidrule(lr){10-13}
\cmidrule(lr){14-17}
& \textbf{MRR} & \textbf{H@1} & \textbf{H@3} & \textbf{H@10}
& \textbf{MRR} & \textbf{H@1} & \textbf{H@3} & \textbf{H@10}
& \textbf{MRR} & \textbf{H@1} & \textbf{H@3} & \textbf{H@10}
& \textbf{MRR} & \textbf{H@1} & \textbf{H@3} & \textbf{H@10} \\
\midrule
LLM-DA \citeyearpar{wang2024large}
& 47.10 & 36.90 & 52.60 & 67.10
& 52.10 & 41.60 & 58.60 & 72.80
& 35.70 & 25.50 & 40.30 & 57.00
& -- & -- & -- & -- \\

MESH \citeyearpar{deng2025multi}
& 44.36 & -- & 49.81 & 64.21
& 48.66 & -- & 54.26 & 68.57
& 33.96 & -- & 38.37 & 54.12
& -- & -- & -- & -- \\

ANRE \citeyearpar{tang2025anre}
& 47.40 & 36.90 & 51.10 & 65.70
& 50.90 & 39.10 & 58.00 & 69.60
& 35.50 & 26.00 & 39.20 & 56.70
& 24.30 & 16.60 & 26.60 & 37.50 \\

TV-LLM \citeyearpar{pan2025leveraging}
& 44.50 & 36.60 & 50.20 & 64.20
& 50.80 & 41.30 & 56.60 & 72.20
& 33.20 & 22.90 & 37.90 & 54.10
& -- & -- & -- & -- \\

CRI \citeyearpar{ning2026critic}
& 49.80 & 38.10 & 55.10 & 70.20
& 53.10 & 40.70 & 59.30 & \textbf{75.10}
& \textbf{38.80} & 27.40 & \textbf{42.10} & 57.80
& -- & -- & -- & -- \\

LANTERN \citeyearpar{jin2026lantern}
& 48.00 & 37.50 & 51.50 & 66.50
& 51.50 & 40.50 & 58.50 & 70.50
& 36.50 & \textbf{28.50} & 40.00 & 57.50
& \textbf{25.50} & 17.50 & 27.00 & 38.00 \\\midrule

\textbf{FreqDiff}
& \textbf{50.85} & \textbf{39.82} & \textbf{57.48} & \textbf{71.48}
& \textbf{54.71} & \textbf{45.02} & \textbf{61.16} & 74.45
& 36.85 & 26.85 & 41.51 & \textbf{61.47}
& 25.04 & \textbf{17.64} & \textbf{27.03} & \textbf{41.83} \\
\bottomrule
\end{tabular}%
}
\end{table*}

As shown in Table \ref{tab:tkg-results}, FreqDiff achieves the best result on 11 of the 16 dataset–metric combinations, including all four metrics on ICEWS14, three metrics on ICEWS05-15, H@10 on ICEWS18, and H@1/H@3/H@10 on GDELT. However, it is not uniformly superior. CRI performs best on ICEWS05-15 H@10 and ICEWS18 MRR/H@3, while LANTERN obtains the strongest ICEWS18 H@1 and GDELT MRR. 

The results also suggest complementary strengths. Diffusion-based FreqDiff is particularly competitive when prediction depends on distributed temporal patterns and when maintaining broad candidate coverage is important, as reflected by its consistent H@10 performance. Its learned denoising representation can aggregate noisy or heterogeneous histories without requiring an explicit symbolic rule to cover each query. LLM-based approaches may be preferable when a query has strong symbolic or semantic support, such as a high-confidence temporal rule, a closely matched historical analogy, or a compact set of highly informative events that can be expressed in the prompt. This is consistent with the design of LLM-DA, TV-LLM, and CRI, which rely on temporal rule induction or validation, and with AnRe and LANTERN, which emphasize analogical demonstrations and carefully selected historical evidence. We provide detailed model descriptions in Appendix \ref{baseline}.

\begin{table}[t]
\centering
\caption{Performance comparison on YAGO and WIKI.}
\label{tab:yago-wiki-results}
\setlength{\tabcolsep}{2pt}
\renewcommand{\arraystretch}{1.1}
\resizebox{\columnwidth}{!}{%
\begin{tabular}{@{}ccccccccc@{}}
\toprule
\multirow{2}{*}{\textbf{Model}}
& \multicolumn{4}{c}{\textbf{YAGO}}
& \multicolumn{4}{c}{\textbf{WIKI}} \\
\cmidrule(lr){2-5}
\cmidrule(lr){6-9}
& \textbf{MRR} & \textbf{H@1} & \textbf{H@3} & \textbf{H@10}
& \textbf{MRR} & \textbf{H@1} & \textbf{H@3} & \textbf{H@10} \\
\midrule
TITer
& 87.47 & 80.09 & 89.96 & 90.27
& 73.91 & 71.70 & 75.41 & 76.93 \\

TiRGN
& 87.95 & 84.34 & 91.37 & 92.92
& 81.65 & 77.77 & 85.12 & 87.08 \\

DiffuTKG
& 88.29 & 84.36 & 91.79 & 93.55
& 82.21 & 78.96 & 85.69 & 88.03 \\

NADEx
& 88.31 & 85.14 & 92.18 & 93.72
& 82.83 & 79.02 & 85.83 & 88.34 \\
\midrule
w/o Freq
& 86.65 & 84.78 & 89.24 & 89.90
& 82.55 & 81.43 & 84.89 & 86.38 \\

w/o $\mathcal{L}_{\mathrm{fft}}$
& 88.74 & 86.62 & 91.87 & 91.23
& 84.08 & 83.03 & 86.92 & 87.99 \\\midrule

\textbf{FreqDiff}
& \textbf{90.32} & \textbf{88.60} & \textbf{93.15} & \textbf{93.91}
& \textbf{85.84} & \textbf{84.76} & \textbf{88.27} & \textbf{89.83} \\
\bottomrule
\end{tabular}}
\end{table}

\subsection{Performance on Knowledge-Centric KGs}
Table \ref{tab:yago-wiki-results} further evaluates FreqDiff on YAGO and WIKI, whose knowledge-centric facts and temporal patterns differ from the event-driven ICEWS and GDELT datasets. FreqDiff consistently outperforms all competing methods across the eight evaluation settings. Compared with the strongest baseline, NADEx, FreqDiff improves MRR and Hit@1 by 2.01 and 3.46 percentage points on YAGO, respectively. The improvements are more pronounced on WIKI, reaching 3.01 points in MRR and 5.74 points in Hit@1. FreqDiff also achieves consistent gains in Hit@3 and Hit@10 on both datasets. These results demonstrate that its effectiveness generalizes beyond geopolitical event forecasting to knowledge-centric temporal graphs.

The ablation results further confirm the contributions of the frequency-aware components. Removing the spectral branch decreases MRR by 3.67 points on YAGO and 3.29 points on WIKI, accompanied by consistent degradation across all Hits metrics. Removing $\mathcal{L}_{\mathrm{fft}}$ also reduces MRR by 1.58 and 1.76 points, respectively. The larger degradation caused by removing the spectral branch highlights the importance of context-aware spectral modeling, while the consistent decline without $\mathcal{L}_{\mathrm{fft}}$ verifies the complementary role of frequency-domain supervision. Together, these results show that both components remain effective across TKGs with distinct temporal characteristics.

\section{Conclusion}
In this paper, we proposed FreqDiff, a frequency-aware diffusion framework for TKG extrapolation. FreqDiff formulates future object prediction as query-slot denoising and reconstructs the target representation with a dual-stream denoiser that combines temporal dependency modeling and context-aware spectral calibration. We further introduced a frequency-domain consistency regularizer to provide explicit spectral supervision for target reconstruction. Experiments on four public TKG benchmarks demonstrate that FreqDiff achieves state-of-the-art performance, while ablation studies verify the effectiveness of spectral calibration and frequency-domain regularization.

\section*{Acknowledgments}
We thank the anonymous reviewers for their valuable discussion and feedback. This work was supported by the National Natural Science Foundation of China (U22B2061).

\section*{Limitation}

This work has two main limitations. First, although we evaluate FreqDiff on four widely used TKG benchmarks, including ICEWS14, ICEWS05–15, ICEWS18, and GDELT, these datasets are primarily centered on political and international relations events. Therefore, the generalizability of FreqDiff to other temporal knowledge graphs, such as those in scientific discovery, financial transactions, public health, or natural disasters, remains to be further examined. Second, FreqDiff introduces context-aware spectral calibration to improve frequency-aware denoising, but its current design relies on a finite set of learnable basis filters. While effective in our experiments, this design may still provide limited flexibility when modeling highly irregular or domain-specific temporal dynamics. Future work may explore more adaptive spectral parameterization strategies and evaluate frequency-aware diffusion reasoning across broader TKG domains.

\section*{Ethics Statement}

This study follows established ethical standards. We use only publicly available benchmark datasets that have been collected and processed by prior research, and our work does not involve new data collection, human-subject interaction, or the use of private personal information. The proposed model is developed for scientific analysis and benchmarking in temporal knowledge graph reasoning. It is not intended for surveillance, deception, profiling, or any harmful application. We respect the data-use terms of the adopted benchmarks and aim to ensure that the research does not compromise the rights, safety, or dignity of any individual or group.

\bibliography{custom}

@data{Boschee15ICEWS,
author = {Boschee, Elizabeth and Lautenschlager, Jennifer and O'Brien, Sean and Shellman, Steve and Starz, James and Ward, Michael},
publisher = {Harvard Dataverse},
title = {{ICEWS Coded Event Data}},
UNF = {UNF:6:NOSHB7wyt0SQ8sMg7+w38w==},
year = {2015},
version = {V37},
doi = {10.7910/DVN/28075},
url = {https://doi.org/10.7910/DVN/28075}
}

@article{Leetaru13gdelt,
  author = {Leetaru, Kalev and Schrodt, Philip A.},
  description = {CiteSeerX — GDELT: Global data on events, location, and tone},
  journal = {ISA Annual Convention},
  title = {GDELT: Global data on events, location, and tone},
 pages = {1–49},
volume = {2},
  url = {http://citeseerx.ist.psu.edu/viewdoc/summary?doi=10.1.1.686.6605},
  year = 2013
}

@inproceedings{Li21Temporal,
author = {Li, Zixuan and Jin, Xiaolong and Li, Wei and Guan, Saiping and Guo, Jiafeng and Shen, Huawei and Wang, Yuanzhuo and Cheng, Xueqi},
title = {Temporal Knowledge Graph Reasoning Based on Evolutional Representation Learning},
year = {2021},
isbn = {9781450380379},
publisher = {Association for Computing Machinery},
address = {New York, NY, USA},
url = {https://doi.org/10.1145/3404835.3462963},
doi = {10.1145/3404835.3462963},
booktitle = {Proceedings of the 44th International ACM SIGIR Conference on Research and Development in Information Retrieval},
pages = {408–417},
numpages = {10},
location = {Virtual Event, Canada},
series = {SIGIR '21}
}

@inproceedings{cai-etal-2024-predicting,
    title = "Predicting the Unpredictable: Uncertainty-Aware Reasoning over Temporal Knowledge Graphs via Diffusion Process",
    author = "Cai, Yuxiang  and
      Liu, Qiao  and
      Gan, Yanglei  and
      Li, Changlin  and
      Liu, Xueyi  and
      Lin, Run  and
      Luo, Da  and 
      Jiaye Yang",
    booktitle = "Findings of the Association for Computational Linguistics: ACL 2024",
    month = aug,
    year = "2024",
    address = "Bangkok, Thailand",
    publisher = "Association for Computational Linguistics",
    url = "https://aclanthology.org/2024.findings-acl.343/",
    doi = "10.18653/v1/2024.findings-acl.343",
    pages = "5766--5778"
}

@inproceedings{chen2024local,
  title={Local-global history-aware contrastive learning for temporal knowledge graph reasoning},
  author={Chen, Wei and Wan, Huaiyu and Wu, Yuting and Zhao, Shuyuan and Cheng, Jiayaqi and Li, Yuxin and Lin, Youfang},
  booktitle={2024 IEEE 40th International Conference on Data Engineering (ICDE)},
  pages={733--746},
  year={2024},
  organization={IEEE}
}

@inproceedings{chen2025llm,
  title={LLM-DR: A Novel LLM-Aided Diffusion Model for Rule Generation on Temporal Knowledge Graphs},
  author={Chen, Kai and Song, Xin and Wang, Ye and Gao, Liqun and Li, Aiping and Zhao, Xiaojuan and Zhou, Bin and Xie, Yalong},
  booktitle={Proceedings of the AAAI Conference on Artificial Intelligence},
  volume={39},
  number={11},
  pages={11481--11489},
  year={2025}
}

@inproceedings{chen2025enhancing,
  title={Enhancing Extrapolation Reasoning on Temporal Knowledge Graphs with Logic Rules and Queries},
  author={Chen, Tingxuan and Yang, Liu and Wang, Zidong and Luo, Shuai and Long, Jun},
  booktitle={ICASSP 2025-2025 IEEE International Conference on Acoustics, Speech and Signal Processing (ICASSP)},
  pages={1--5},
  year={2025},
  organization={IEEE}
}

@inproceedings{chen2025cogntke,
  title={CognTKE: A Cognitive Temporal Knowledge Extrapolation Framework},
  author={Chen, Wei and Wu, Yuting and Wu, Shuhan and Zhang, Zhiyu and Liao, Mengqi and Lin, Youfang and Wan, Huaiyu},
  booktitle={Proceedings of the AAAI Conference on Artificial Intelligence},
  volume={39},
  number={14},
  pages={14815--14823},
  year={2025}
}

@inproceedings{garcia2018learning,
  title={Learning Sequence Encoders for Temporal Knowledge Graph Completion},
  author={Garcia-Duran, Alberto and Duman{\v{c}}i{\'c}, Sebastijan and Niepert, Mathias},
  booktitle={Proceedings of the 2018 Conference on Empirical Methods in Natural Language Processing},
  pages={4816--4821},
  year={2018}
}

@inproceedings{leblay2018deriving,
  title={Deriving validity time in knowledge graph},
  author={Leblay, Julien and Chekol, Melisachew Wudage},
  booktitle={Companion proceedings of the the web conference 2018},
  pages={1771--1776},
  year={2018}
}

@inproceedings{yang2015embedding,
  title={Embedding Entities and Relations for Learning and Inference in Knowledge Bases},
  author={Yang, Bishan and Yih, Scott Wen-tau and He, Xiaodong and Gao, Jianfeng and Deng, Li},
  booktitle={Proceedings of the International Conference on Learning Representations (ICLR) 2015},
  year={2015}
}

@inproceedings{dettmers2018convolutional,
  title={Convolutional 2d knowledge graph embeddings},
  author={Dettmers, Tim and Minervini, Pasquale and Stenetorp, Pontus and Riedel, Sebastian},
  booktitle={Proceedings of the AAAI conference on artificial intelligence},
  volume={32},
  number={1},
  year={2018}
}

@inproceedings{sun2018rotate,
title={RotatE: Knowledge Graph Embedding by Relational Rotation in Complex Space},
author={Zhiqing Sun and Zhi-Hong Deng and Jian-Yun Nie and Jian Tang},
booktitle={International Conference on Learning Representations},
year={2019},
url={https://openreview.net/forum?id=HkgEQnRqYQ},
}

@inproceedings{goel2020diachronic,
  title={Diachronic embedding for temporal knowledge graph completion},
  author={Goel, Rishab and Kazemi, Seyed Mehran and Brubaker, Marcus and Poupart, Pascal},
  booktitle={Proceedings of the AAAI conference on artificial intelligence},
  volume={34},
  number={04},
  pages={3988--3995},
  year={2020}
}

@inproceedings{jin2020recurrent,
  title={Recurrent Event Network: Autoregressive Structure Inferenceover Temporal Knowledge Graphs},
  author={Jin, Woojeong and Qu, Meng and Jin, Xisen and Ren, Xiang},
  booktitle={Proceedings of the 2020 Conference on Empirical Methods in Natural Language Processing (EMNLP)},
  pages={6669--6683},
  year={2020}
}

@inproceedings{zhu2021learning,
  title={Learning from history: Modeling temporal knowledge graphs with sequential copy-generation networks},
  author={Zhu, Cunchao and Chen, Muhao and Fan, Changjun and Cheng, Guangquan and Zhang, Yan},
  booktitle={Proceedings of the AAAI conference on artificial intelligence},
  volume={35},
  number={5},
  pages={4732--4740},
  year={2021}
}

@inproceedings{li2022complex,
  title={Complex Evolutional Pattern Learning for Temporal Knowledge Graph Reasoning},
  author={Li, Zixuan and Guan, Saiping and Jin, Xiaolong and Peng, Weihua and Lyu, Yajuan and Zhu, Yong and Bai, Long and Li, Wei and Guo, Jiafeng and Cheng, Xueqi},
  booktitle={Proceedings of the 60th Annual Meeting of the Association for Computational Linguistics (Volume 2: Short Papers)},
  pages={290--296},
  year={2022}
}

@inproceedings{li2022tirgn,
  title={TiRGN: Time-Guided Recurrent Graph Network with Local-Global Historical Patterns for Temporal Knowledge Graph Reasoning.},
  author={Li, Yujia and Sun, Shiliang and Zhao, Jing},
  booktitle={IJCAI},
  pages={2152--2158},
  year={2022}
}

@inproceedings{li2022hismatch,
  title={HiSMatch: Historical Structure Matching based Temporal Knowledge Graph Reasoning},
  author={Li, Zixuan and Hou, Zhongni and Guan, Saiping and Jin, Xiaolong and Peng, Weihua and Bai, Long and Lyu, Yajuan and Li, Wei and Guo, Jiafeng and Cheng, Xueqi},
  booktitle={Findings of the Association for Computational Linguistics: EMNLP 2022},
  pages={7328--7338},
  year={2022}
}

@inproceedings{liu2023retia,
  title={RETIA: relation-entity twin-interact aggregation for temporal knowledge graph extrapolation},
  author={Liu, Kangzheng and Zhao, Feng and Xu, Guandong and Wang, Xianzhi and Jin, Hai},
  booktitle={2023 IEEE 39th international conference on data engineering (ICDE)},
  pages={1761--1774},
  year={2023},
  organization={IEEE}
}

@inproceedings{xu2023temporal,
  title={Temporal knowledge graph reasoning with historical contrastive learning},
  author={Xu, Yi and Ou, Junjie and Xu, Hui and Fu, Luoyi},
  booktitle={Proceedings of the AAAI conference on artificial intelligence},
  volume={37},
  number={4},
  pages={4765--4773},
  year={2023}
}

@article{chen2024thcn,
  title={THCN: A Hawkes Process Based Temporal Causal Convolutional Network for Extrapolation Reasoning in Temporal Knowledge Graphs},
  author={Chen, Tingxuan and Long, Jun and Wang, Zidong and Luo, Shuai and Huang, Jincai and Yang, Liu},
  journal={IEEE Transactions on Knowledge and Data Engineering},
  year={2024},
  publisher={IEEE}
}

@article{liang2024survey,
  title={A survey of knowledge graph reasoning on graph types: Static, dynamic, and multi-modal},
  author={Liang, Ke and Meng, Lingyuan and Liu, Meng and Liu, Yue and Tu, Wenxuan and Wang, Siwei and Zhou, Sihang and Liu, Xinwang and Sun, Fuchun and He, Kunlun},
  journal={IEEE Transactions on Pattern Analysis and Machine Intelligence},
  year={2024},
  publisher={IEEE}
}

@article{ji2021survey,
  title={A survey on knowledge graphs: Representation, acquisition, and applications},
  author={Ji, Shaoxiong and Pan, Shirui and Cambria, Erik and Marttinen, Pekka and Philip, S Yu},
  journal={IEEE transactions on neural networks and learning systems},
  volume={33},
  number={2},
  pages={494--514},
  year={2021},
  publisher={IEEE}
}

@inproceedings{cai2023temporal,
  title={Temporal knowledge graph completion: a survey},
  author={Cai, Borui and Xiang, Yong and Gao, Longxiang and Zhang, He and Li, Yunfeng and Li, Jianxin},
  booktitle={Proceedings of the Thirty-Second International Joint Conference on Artificial Intelligence},
  pages={6545--6553},
  year={2023}
}

@inproceedings{trivedi2017know,
  title={Know-evolve: Deep temporal reasoning for dynamic knowledge graphs},
  author={Trivedi, Rakshit and Dai, Hanjun and Wang, Yichen and Song, Le},
  booktitle={international conference on machine learning},
  pages={3462--3471},
  year={2017},
  organization={PMLR}
}

@article{bordes2013translating,
  title={Translating embeddings for modeling multi-relational data},
  author={Bordes, Antoine and Usunier, Nicolas and Garcia-Duran, Alberto and Weston, Jason and Yakhnenko, Oksana},
  journal={Advances in neural information processing systems},
  volume={26},
  year={2013}
}

@inproceedings{zhang2023learning,
  title={Learning long-and short-term representations for temporal knowledge graph reasoning},
  author={Zhang, Mengqi and Xia, Yuwei and Liu, Qiang and Wu, Shu and Wang, Liang},
  booktitle={Proceedings of the ACM web conference 2023},
  pages={2412--2422},
  year={2023}
}

@inproceedings{sun2021timetraveler,
  title={TimeTraveler: Reinforcement Learning for Temporal Knowledge Graph Forecasting},
  author={Sun, Haohai and Zhong, Jialun and Ma, Yunpu and Han, Zhen and He, Kun},
  booktitle={Proceedings of the 2021 Conference on Empirical Methods in Natural Language Processing},
  pages={8306--8319},
  year={2021}
}

@inproceedings{cao2025dpcl,
  title={DPCL-Diff: Temporal Knowledge Graph Reasoning Based on Graph Node Diffusion Model with Dual-Domain Periodic Contrastive Learning},
  author={Cao, Yukun and Wang, Lisheng and Huang, Luobin},
  booktitle={Proceedings of the AAAI Conference on Artificial Intelligence},
  volume={39},
  number={14},
  pages={14806--14814},
  year={2025}
}

@inproceedings{han2021explainable,
title={Explainable Subgraph Reasoning for Forecasting on Temporal Knowledge Graphs},
author={Zhen Han and Peng Chen and Yunpu Ma and Volker Tresp},
booktitle={International Conference on Learning Representations},
year={2021},
url={https://openreview.net/forum?id=pGIHq1m7PU}
}

@inproceedings{liu2022tlogic,
  title={Tlogic: Temporal logical rules for explainable link forecasting on temporal knowledge graphs},
  author={Liu, Yushan and Ma, Yunpu and Hildebrandt, Marcel and Joblin, Mitchell and Tresp, Volker},
  booktitle={Proceedings of the AAAI conference on artificial intelligence},
  volume={36},
  number={4},
  pages={4120--4127},
  year={2022}
}

@inproceedings{dong2023adaptive,
  title={Adaptive path-memory network for temporal knowledge graph reasoning},
  author={Dong, Hao and Ning, Zhiyuan and Wang, Pengyang and Qiao, Ziyue and Wang, Pengfei and Zhou, Yuanchun and Fu, Yanjie},
  booktitle={Proceedings of the Thirty-Second International Joint Conference on Artificial Intelligence},
  pages={2086--2094},
  year={2023}
}

@inproceedings{sohl2015deep,
  title={Deep unsupervised learning using nonequilibrium thermodynamics},
  author={Sohl-Dickstein, Jascha and Weiss, Eric and Maheswaranathan, Niru and Ganguli, Surya},
  booktitle={International conference on machine learning},
  pages={2256--2265},
  year={2015},
  organization={pmlr}
}

@article{dhariwal2021diffusion,
  title={Diffusion models beat gans on image synthesis},
  author={Dhariwal, Prafulla and Nichol, Alexander},
  journal={Advances in neural information processing systems},
  volume={34},
  pages={8780--8794},
  year={2021}
}

@inproceedings{nichol2022glide,
  title={GLIDE: Towards Photorealistic Image Generation and Editing with Text-Guided Diffusion Models},
  author={Nichol, Alexander Quinn and Dhariwal, Prafulla and Ramesh, Aditya and Shyam, Pranav and Mishkin, Pamela and Mcgrew, Bob and Sutskever, Ilya and Chen, Mark},
  booktitle={International Conference on Machine Learning},
  pages={16784--16804},
  year={2022},
  organization={PMLR}
}

@inproceedings{kongdiffwave,
  title={DiffWave: A Versatile Diffusion Model for Audio Synthesis},
  author={Kong, Zhifeng and Ping, Wei and Huang, Jiaji and Zhao, Kexin and Catanzaro, Bryan},
  booktitle={International Conference on Learning Representations},
  year={2020}
}

@inproceedings{liu2023audioldm,
  title={AudioLDM: Text-to-Audio Generation with Latent Diffusion Models},
  author={Liu, Haohe and Chen, Zehua and Yuan, Yi and Mei, Xinhao and Liu, Xubo and Mandic, Danilo and Wang, Wenwu and Plumbley, Mark D},
  booktitle={International Conference on Machine Learning},
  pages={21450--21474},
  year={2023},
  organization={PMLR}
}

@article{li2022diffusion,
  title={Diffusion-lm improves controllable text generation},
  author={Li, Xiang and Thickstun, John and Gulrajani, Ishaan and Liang, Percy S and Hashimoto, Tatsunori B},
  journal={Advances in neural information processing systems},
  volume={35},
  pages={4328--4343},
  year={2022}
}

@inproceedings{gongdiffuseq,
  title={DiffuSeq: Sequence to Sequence Text Generation with Diffusion Models},
  author={Gong, Shansan and Li, Mukai and Feng, Jiangtao and Wu, Zhiyong and Kong, Lingpeng},
  booktitle={The Eleventh International Conference on Learning Representations},
  year={2022}
}

@inproceedings{gong2023diffuseq,
  title={DiffuSeq-v2: Bridging Discrete and Continuous Text Spaces for Accelerated Seq2Seq Diffusion Models},
  author={Gong, Shansan and Li, Mukai and Feng, Jiangtao and Wu, Zhiyong and Kong, Lingpeng},
  booktitle={Findings of the Association for Computational Linguistics: EMNLP 2023},
  pages={9868--9875},
  year={2023}
}

@article{yang2023generate,
  title={Generate what you prefer: Reshaping sequential recommendation via guided diffusion},
  author={Yang, Zhengyi and Wu, Jiancan and Wang, Zhicai and Wang, Xiang and Yuan, Yancheng and He, Xiangnan},
  journal={Advances in Neural Information Processing Systems},
  volume={36},
  pages={24247--24261},
  year={2023}
}

@inproceedings{shen2023diffusionner,
  title={DiffusionNER: Boundary Diffusion for Named Entity Recognition},
  author={Shen, Yongliang and Song, Kaitao and Tan, Xu and Li, Dongsheng and Lu, Weiming and Zhuang, Yueting},
  booktitle={Proceedings of the 61st Annual Meeting of the Association for Computational Linguistics (Volume 1: Long Papers)},
  pages={3875--3890},
  year={2023}
}

@inproceedings{wang2023diffusion,
  title={Diffusion recommender model},
  author={Wang, Wenjie and Xu, Yiyan and Feng, Fuli and Lin, Xinyu and He, Xiangnan and Chua, Tat-Seng},
  booktitle={Proceedings of the 46th International ACM SIGIR Conference on Research and Development in Information Retrieval},
  pages={832--841},
  year={2023}
}

@article{kazemi2018simple,
  title={Simple embedding for link prediction in knowledge graphs},
  author={Kazemi, Seyed Mehran and Poole, David},
  journal={Advances in neural information processing systems},
  volume={31},
  year={2018}
}

@article{wang2024large,
  title={Large language models-guided dynamic adaptation for temporal knowledge graph reasoning},
  author={Wang, Jiapu and Kai, Sun and Luo, Linhao and Wei, Wei and Hu, Yongli and Liew, Alan Wee-Chung and Pan, Shirui and Yin, Baocai},
  journal={Advances in Neural Information Processing Systems},
  volume={37},
  pages={8384--8410},
  year={2024}
}

@article{luo2024chain,
  title={Chain of History: Learning and Forecasting with LLMs for Temporal Knowledge Graph Completion},
  author={Luo, Ruilin and Gu, Tianle and Li, Haoling and Li, Junzhe and Lin, Zicheng and Li, Jiayi and Yang, Yujiu},
  journal={CoRR},
  year={2024}
}

@inproceedings{gan2026negative,
  title={Negative-Aware Diffusion Process for Temporal Knowledge Graph Extrapolation},
  author={Gan, Yanglei and He, Peng and Cai, Yuxiang and Lin, Run and Zhou, Guanyu and Liu, Qiao},
  booktitle={Findings of the Association for Computational Linguistics: EACL 2026},
  pages={3352--3367},
  year={2026}
}

@article{duhamel1990fast,
  title={Fast Fourier transforms: a tutorial review and a state of the art},
  author={Duhamel, Pierre and Vetterli, Martin},
  journal={Signal processing},
  volume={19},
  number={4},
  pages={259--299},
  year={1990},
  publisher={Elsevier}
}

@article{frigo2005design,
  title={The design and implementation of FFTW3},
  author={Frigo, Matteo and Johnson, Steven G},
  journal={Proceedings of the IEEE},
  volume={93},
  number={2},
  pages={216--231},
  year={2005},
  publisher={IEEE}
}

@inproceedings{chen2025frequency,
  title={Frequency-dynamic attention modulation for dense prediction},
  author={Chen, Linwei and Gu, Lin and Fu, Ying},
  booktitle={Proceedings of the IEEE/CVF International Conference on Computer Vision},
  pages={22620--22632},
  year={2025}
}

@inproceedings{cai2021frequency,
  title={Frequency domain image translation: More photo-realistic, better identity-preserving},
  author={Cai, Mu and Zhang, Hong and Huang, Huijuan and Geng, Qichuan and Li, Yixuan and Huang, Gao},
  booktitle={Proceedings of the IEEE/CVF International Conference on Computer Vision},
  pages={13930--13940},
  year={2021}
}

@inproceedings{tatsunami2024fft,
  title={Fft-based dynamic token mixer for vision},
  author={Tatsunami, Yuki and Taki, Masato},
  booktitle={Proceedings of the AAAI Conference on Artificial Intelligence},
  volume={38},
  number={14},
  pages={15328--15336},
  year={2024}
}

@article{tamkin2020language,
  title={Language through a prism: A spectral approach for multiscale language representations},
  author={Tamkin, Alex and Jurafsky, Dan and Goodman, Noah},
  journal={Advances in Neural Information Processing Systems},
  volume={33},
  pages={5492--5504},
  year={2020}
}

@inproceedings{lee2022fnet,
  title={FNet: Mixing Tokens with Fourier Transforms},
  author={Lee-Thorp, James and Ainslie, Joshua and Eckstein, Ilya and Ontanon, Santiago},
  booktitle={Proceedings of the 2022 Conference of the North American Chapter of the Association for Computational Linguistics: Human Language Technologies},
  pages={4296--4313},
  year={2022}
}

@inproceedings{wang2025filterts,
  title={Filterts: Comprehensive frequency filtering for multivariate time series forecasting},
  author={Wang, Yulong and Liu, Yushuo and Duan, Xiaoyi and Wang, Kai},
  booktitle={Proceedings of the AAAI Conference on Artificial Intelligence},
  volume={39},
  number={20},
  pages={21375--21383},
  year={2025}
}

@inproceedings{xu2020tero,
  title={TeRo: A time-aware knowledge graph embedding via temporal rotation},
  author={Xu, Chengjin and Nayyeri, Mojtaba and Alkhoury, Fouad and Yazdi, Hamed Shariat and Lehmann, Jens},
  booktitle={Proceedings of the 28th International Conference on Computational Linguistics},
  pages={1583--1593},
  year={2020}
}

@inproceedings{sadeghian2021chronor,
  title={Chronor: Rotation based temporal knowledge graph embedding},
  author={Sadeghian, Ali and Armandpour, Mohammadreza and Colas, Anthony and Wang, Daisy Zhe},
  booktitle={Proceedings of the AAAI conference on artificial intelligence},
  volume={35},
  number={7},
  pages={6471--6479},
  year={2021}
}

@inproceedings{li2023teast,
  title={Teast: Temporal knowledge graph embedding via archimedean spiral timeline},
  author={Li, Jiang and Su, Xiangdong and Gao, Guanglai},
  booktitle={Proceedings of the 61st Annual Meeting of the Association for Computational Linguistics (Volume 1: Long Papers)},
  pages={15460--15474},
  year={2023}
}

@inproceedings{liu2025terdy,
  title={Terdy: Temporal relation dynamics through frequency decomposition for temporal knowledge graph completion},
  author={Liu, Ziyang and Wang, Chaokun},
  booktitle={Proceedings of the 63rd Annual Meeting of the Association for Computational Linguistics (Volume 1: Long Papers)},
  pages={9611--9622},
  year={2025}
}

@inproceedings{chen2024natural,
  title={Natural evolution-based dual-level aggregation for temporal knowledge graph reasoning},
  author={Chen, Bin and Xiao, Chunjing and Zhou, Fan},
  booktitle={Findings of the association for computational linguistics: EMNLP 2024},
  pages={9274--9284},
  year={2024}
}

@inproceedings{zhang2023learn,
  title={Learning latent relations for temporal knowledge graph reasoning},
  author={Zhang, Mengqi and Xia, Yuwei and Liu, Qiang and Wu, Shu and Wang, Liang},
  booktitle={Proceedings of the 61st Annual Meeting of the Association for Computational Linguistics (Volume 1: Long Papers)},
  pages={12617--12631},
  year={2023}
}

@inproceedings{wang2025dltkg,
    title = "{DLTKG}: Denoising Logic-based Temporal Knowledge Graph Reasoning",
    author = "Wang, Xiaoke  and
      Zhang, Fu  and
      Cheng, Jingwei  and
      Chi, Yiwen  and
      Peng, Jiashun  and
      Ning, Yingsong",
    booktitle = "Findings of the Association for Computational Linguistics: EMNLP 2025",
    month = nov,
    year = "2025",
    address = "Suzhou, China",
    publisher = "Association for Computational Linguistics",
    url = "https://aclanthology.org/2025.findings-emnlp.1017/",
    doi = "10.18653/v1/2025.findings-emnlp.1017",
    pages = "18730--18743",
    ISBN = "979-8-89176-335-7",
}

@inproceedings{he2026fait,
  title={FAiT: Frequency-Aware Inverted Transformer for Multivariate Time Series Forecasting},
  author={He, Peng and Liu, Yao and Gan, Yanglei and Lin, Run and Cai, Yuxiang and Liu, Qiao},
  booktitle={Proceedings of the 32nd ACM SIGKDD Conference on Knowledge Discovery and Data Mining V. 2},
  pages={1614--1625},
  year={2026}
}

@inproceedings{he2026exploiting,
  title={Exploiting inter-session information with frequency-enhanced dual-path networks for sequential recommendation},
  author={He, Peng and Gan, Yanglei and Dai, Tingting and Lin, Run and Li, Xuexin and Liu, Yao and Liu, Qiao},
  booktitle={Proceedings of the AAAI Conference on Artificial Intelligence},
  volume={40},
  number={17},
  pages={14820--14828},
  year={2026}
}

@inproceedings{deng2025multi,
  title={A multi-expert structural-semantic hybrid framework for unveiling historical patterns in temporal knowledge graphs},
  author={Deng, Yimin and Wu, Yuxia and Wang, Yejing and Zhao, Guoshuai and Zhu, Li and Liu, Qidong and Xu, Derong and Fu, Zichuan and Wu, Xian and Zheng, Yefeng and others},
  booktitle={Findings of the Association for Computational Linguistics: ACL 2025},
  pages={20553--20565},
  year={2025}
}

@inproceedings{tang2025anre,
  title={AnRe: Analogical replay for temporal knowledge graph forecasting},
  author={Tang, Guo and Chu, Zheng and Zheng, Wenxiang and Xiang, Junjia and Li, Yizhuo and Zhang, Weihao and Liu, Ming and Qin, Bing},
  booktitle={Proceedings of the 63rd Annual Meeting of the Association for Computational Linguistics (Volume 1: Long Papers)},
  pages={4632--4650},
  year={2025}
}

@article{pan2025leveraging,
  title={Leveraging temporal validity of rules via LLMs for enhanced temporal knowledge graph reasoning},
  author={Pan, Qihong and Yao, Limin and Shen, Guojiang and Han, Xiao and Chen, Yichuan and Kong, Xiangjie},
  journal={Knowledge-based systems},
  pages={114094},
  year={2025},
  publisher={Elsevier}
}

@inproceedings{ning2026critic,
  title={Critic Rule Induction: Improving Temporal Knowledge Graph Forecasting with Generator-Critic Language Models},
  author={Ning, Yingsong and Zhang, Fu and Cheng, Jingwei and Peng, Jiashun and Wang, Xiaoke},
  booktitle={Findings of the Association for Computational Linguistics: ACL 2026},
  pages={29436--29448},
  year={2026}
}

@inproceedings{jin2026lantern,
  title={LANTERN in the Event Stream: Training-Free Temporal Knowledge Graph Forecasting by Balancing Inertia and Shifts},
  author={Jin, Chengyuan and Chang, Ao and Zeng, Daojian and Teng, Wenhao and Liao, Xiangwen and Liu, Kang and Zhao, Jun and Chen, Yubo},
  booktitle={Findings of the Association for Computational Linguistics: ACL 2026},
  pages={11519--11533},
  year={2026}
}

\newpage

\appendix

\section{Additional Related Work}\label{sec:a}

\subsection{Spectral Representation Learning}
Spectral analysis, most commonly implemented with the Discrete Fourier Transform (DFT), decomposes signals into frequency components for efficient processing \cite{duhamel1990fast,frigo2005design}. Motivated by the convolution theorem, recent deep models incorporate spectral transforms to capture global dependencies and improve efficiency. This idea has been adopted in computer vision \cite{cai2021frequency,tatsunami2024fft,chen2025frequency}, natural language processing \cite{tamkin2020language,lee2022fnet}, sequential recommendation \cite{he2026exploiting}, and time-series forecasting \cite{wang2025filterts,he2026fait}. 

In the realm of Temporal Knowledge Graph (TKG) reasoning, explicit spectral learning remains relatively underexplored, although several earlier TKG Completion (TKGC) models can be regarded as important precursors. Diachronic Embedding introduces sine-based temporal entity functions, and its expressivity analysis explicitly relates this parameterization to Fourier sine series \cite{goel2020diachronic}. Along a related line, TeRo \cite{xu2020tero} and ChronoR \cite{sadeghian2021chronor} model temporal evolution through rotations in complex or high-dimensional embedding spaces, while TeAST \cite{li2023teast} maps relations onto an Archimedean spiral timeline. These methods capture phase-sensitive, periodic, or geometrically regular temporal patterns, but they mainly encode temporal regularity through parametric embedding functions rather than explicitly transforming representations into the frequency domain. More recently, TeRDy \cite{liu2025terdy} establishes a more direct connection to spectral learning by applying FFT-based low-pass and high-pass decomposition to relation embeddings, thereby separating long-term and short-term temporal relation dynamics.

\subsection{Diffusion Models on Discrete Data}
Diffusion models (DMs) \cite{sohl2015deep} have become a powerful generative paradigm, achieving strong performance in image generation \cite{dhariwal2021diffusion,nichol2022glide} and audio synthesis \cite{kongdiffwave,liu2023audioldm}. Although early diffusion models were mainly designed for continuous Euclidean spaces, recent studies have extended them to discrete symbolic data. For text generation, Diffusion-LM \cite{li2022diffusion} maps word tokens into continuous embeddings and performs denoising in the latent space, while DiffuSeq \cite{gongdiffuseq,gong2023diffuseq} introduces a sequence-level corruption and denoising process to support coherent non-autoregressive generation. Diffusion has also been adapted to structured prediction tasks. DiffusionNER \cite{shen2023diffusionner} formulates named entity recognition as a span-boundary denoising problem, gradually refining noisy boundaries into valid entity predictions.

Beyond text and structured prediction, diffusion models have been applied to symbolic interaction modeling. In recommendation, DiffRec \cite{wang2023diffusion} and DreamRec \cite{yang2023generate} inject noise into user–item interaction histories and learn the reverse process to capture uncertain preference distributions. These developments suggest that diffusion is not limited to dense continuous signals, but can also provide a flexible generative mechanism for discrete data, where iterative denoising helps model latent structure, uncertainty, and complex dependencies.

\section{Preliminary}

\noindent\textbf{Definition 3. Discrete Fourier Transform. } The Discrete Fourier Transform (DFT) is a fundamental tool in digital signal processing. Given a length $N$ time-domain sequence $x[n]$, the DFT maps it to the frequency domain via:
\begin{equation}
\small
    \mathcal{X}[k]=\sum_{n=0}^{N-1}x[n]e^{-j2\pi kn/N}, \quad k=0,1,...,N-1,
\end{equation}
where $j$ is the imaginary unit and $\mathcal{X}[k]$ is the complex spectral coefficient associated with the discrete frequency $\omega_k = 2\pi k/N$. Each $\mathcal{X}[k]$ can be decomposed into real and imaginary parts:
\begin{equation}
\small
\begin{aligned} 
    \mathcal{X}[k] = \textbf{Real}(\mathcal{X}[k]) + j\textbf{Imag}(\mathcal{X}[k]), \\
    \textbf{Real}(\mathcal{X}[k])=\sum_{n=0}^{N-1}x[n]\cos\left(\frac{2\pi}{N}kn\right),\\ \textbf{Imag}(\mathcal{X}[k])=-\sum_{n=0}^{N-1}x[n]\sin\left(\frac{2\pi}{N}kn\right). 
\end{aligned}
\end{equation}
The inverse DFT (IDFT) reconstructs the original sequence via:
\begin{equation}
\small
x[n]=\frac{1}{N}\sum_{k=0}^{N-1}\mathcal{X}[k]e^{j2\pi kn/N},\quad n=0,1,\dots,N-1.
\end{equation}
In short, we denote DFT and IDFT operators as $\mathcal{F}$, $\mathcal{F}^{-1}$, respectively.

\section{Experimental Setup}

\subsection{Dataset Statistics} \label{data stat}
To ensure consistency and comparability with prior TKG extrapolation studies, we follow the chronological splitting protocol adopted by \cite{Li21Temporal,cai-etal-2024-predicting,gan2026negative}, where the earliest 80\% of facts are used for training, the subsequent 10\% for validation, and the latest 10\% for testing. This protocol preserves the temporal order of observed facts and avoids information leakage from future timestamps during model training.

\begin{table}[t]
\setlength{\tabcolsep}{1.5pt} 
\fontsize{8pt}{12pt}\selectfont 
\caption{The statistics of the datasets. $|E|$ and $|R|$ denote the number of unique entities and event types, respectively.}
\begin{tabular}{@{}ccccccc@{}}
\toprule
Datasets   & $|E|$ & $|R|$    & \textit{Train}  & \textit{Valid} & \textit{Test} & \textit{Unseen Ratio} \\ \midrule
ICEWS14    & 6,869  & 230 & 74,845    & 8,514   & 7,371   & 58.43\% \\
ICEWS18    & 23,033 & 256 & 373,018   & 45,995  & 49,545  & 55.69\% \\
ICEWS05-15 & 10,094 & 251 & 368,868   & 46,302  & 46,159  & 39.82\% \\
GDELT & 7,691  & 240 & 1,734,399 & 238,765 & 305,241 & 43.72\%  \\ \bottomrule
\end{tabular}\label{stat}
\end{table}

\begin{table}[]
\centering
\small
\caption{Implementation details for each benchmark.}
\label{tab:filterbank_implementation_details}
\resizebox{\linewidth}{!}{%
\begin{tabular}{ccccc}
\toprule
\textbf{Hyperparameter} & \textbf{ICEWS14} & \textbf{ICEWS18} & \textbf{GDELT} & \textbf{ICEWS05-15} \\
\midrule
Hidden dimension & 200 & 200 & 200 & 200 \\
Dropout & 0.2 & 0.2 & 0.2 & 0.2 \\
Embedding dropout & 0.2 & 0.2 & 0.2 & 0.2 \\
Maximum history length & 128 & 128 & 128 & 128 \\
Maximum input length & 64 & 64 & 64 & 64 \\
Number of epochs & 100 & 100 & 100 & 100 \\
Learning rate & $1e^{-3}$ & $1e^{-3}$ & $5e^{-4}$ & $5e^{-4}$ \\
Diffusion steps & 200 & 200 & 200 & 200 \\
$\alpha$ & 0.7 & 0.7 & 0.7 & 0.7 \\
$G$ & 8 & 10 & 4 & 8 \\
$F$ & 20 & 8 & 8 & 10 \\
$\lambda$ & 0.3 & 0.3 & 0.3 & 0.3 \\
FFT loss type & L1 & L1 & L1 & L1 \\
\bottomrule
\end{tabular}%
}
\end{table}

\subsection{Implementation Details} \label{implemt}
Table \ref{tab:filterbank_implementation_details} summarizes the main implementation settings. For all datasets, the hidden dimension is set to 200, with both dropout and embedding dropout fixed at 0.2. The maximum history length and input length are set to 128 and 64, respectively, and all models are trained for 100 epochs with 200 diffusion steps. Most hyperparameters are kept consistent across datasets to ensure fair comparison, while the learning rate and filterbank configuration are adjusted according to dataset characteristics. Specifically, ICEWS14 and ICEWS18 use a learning rate of $1e^{-3}$, whereas GDELT and ICEWS05-15 adopt a smaller learning rate of $5e^{-4}$. The fusion weight $\alpha$ and regularization weight $\lambda$ are fixed at 0.7 and 0.3, respectively, and the FFT regularization adopts an L1 loss for all datasets.

\subsection{Baselines} \label{baseline}

\paragraph{Static Baselines:}
\begin{itemize}[leftmargin=*, itemsep=0pt, parsep=0pt]
    \item DistMult \cite{yang2015embedding}, employs a bilinear scoring function, modeling triple plausibility via a relation matrix that captures pairwise interactions between subject and object embeddings.
    \item ConvE \cite{dettmers2018convolutional}, applies 2D convolution over reshaped entity and relation embeddings, followed by a projection layer to learn richer feature interactions for link prediction.
    \item RotatE \cite{sun2018rotate}, represents relations as complex-valued rotations in the embedding space, enabling the model to naturally encode and infer diverse relational patterns such as symmetry and inversion.
\end{itemize}

\paragraph{Interpolation Baselines:}

\begin{itemize}[leftmargin=*, itemsep=0pt, parsep=0pt]
    \item TTransE \cite{leblay2018deriving}, explicitly models temporal dynamics by embedding entities and relations within a continuous time framework, using translation operations along the time dimension to capture their evolution.
    \item TA-DistMult \cite{garcia2018learning}, extends the DistMult scoring function with time‐aware embeddings, allowing the model to adapt relation parameters according to temporal context.
    \item DE-SimplE \cite{goel2020diachronic}, employs diachronic embeddings that parameterize entity and relation representations as functions of time, hence modeling progressive change across different timestamps.
\end{itemize}

 

\paragraph{Extrapolation Baselines:}

\begin{itemize}[leftmargin=*, itemsep=0pt, parsep=0pt]
    \item RE-NET \cite{jin2020recurrent}, integrates recurrent neural architectures with graph convolution to capture the sequential evolution of entities and predict future links.
    \item Re-GCN \cite{Li21Temporal}, employs a Recurrent Evolutionary GCN that recurrently updates entity and relation embeddings at each timestamp by propagating temporal signals through the KG.
    \item CyGNet \cite{zhu2021learning}, models cyclical temporal patterns, enabling the learning of periodic behaviors and extrapolate yet-unseen links.
    \item TITer \cite{sun2021timetraveler}, applies hierarchical transformations to entity embeddings, iteratively tracking their evolution to anticipate future TKG states.
    \item CEN \cite{li2022complex}, uses length-aware convolutional filters to extract multi-scale evolutionary patterns, with an online training strategy to handle temporal variability.
    \item TiRGN \cite{li2022tirgn}, leverages recurrent graph networks to encode dynamic relational structures, improving inference of unseen facts.
    \item RETIA \cite{liu2023retia}, constructs a twin hyper-relation subgraph and evolutionarily aggregates adjacent entity and relation features for enriched message passing.
    \item CENET \cite{xu2023temporal}, incorporates contrastive learning objectives to strengthen dynamic representation learning within temporal graphs.
    \item THCN \cite{chen2024thcn}, introduces temporal causal convolutional networks grounded in Hawkes processes to distinguish the relative importance of concurrent facts.
    \item DiffuTKG \cite{cai-etal-2024-predicting}, reframes TKG reasoning as a denoising diffusion process over entity embeddings to generate future links.
    \item LogiQ \cite{chen2025enhancing}, augments diffusion-based generation with logical constraints, improving both accuracy and interpretability.
    \item CognTKE \cite{chen2025cogntke}, integrates cognitively symbolic priors into embedding, enabling more transparent and reliable extrapolation.
    \item NADEx \cite{gan2026negative}, incorporates negative-aware diffusion to better distinguish plausible future facts from spurious candidates.
\end{itemize}

\paragraph{LLM-based Forecasters:}

\begin{itemize}[leftmargin=*, itemsep=0pt, parsep=0pt]
\item LLM-DA \cite{wang2024large}, generates temporal rules using LLMs and dynamically updates them based on recent events.
\item MESH \cite{deng2025multi}, employs multiple expert modules to integrate structural and semantic information for future link prediction.
\item AnRe \cite{tang2025anre}, combines long- and short-term historical contexts with LLM-generated analogical demonstrations to support temporal reasoning.
\item TV-LLM \cite{pan2025leveraging}, models the validity of LLM-generated rules and integrates rule-based retrieval with graph-based candidate scoring.
\item CRI \cite{ning2026critic}, adopts a generator-critic framework that uses fact-grounded evaluation to validate LLM-induced rules before reasoning.
\item LANTERN \cite{jin2026lantern}, constructs reasoning prompts by balancing long-term interaction strength and short-term novelty, complemented by structure-aware analogical demonstrations.
\end{itemize}

\begin{figure*}[t]
	\centering
	\subfloat[FreqDiff w/o. Freq]{
		\includegraphics[scale=0.45]{viz/ICEWS14_woFreq.pdf}
	}%
 	\subfloat[FreqDiff]{
		\includegraphics[scale=0.45]{viz/ICEWS14_wFreq.pdf}
	}%
 	\subfloat[DiffuTKG]{
		\includegraphics[scale=0.45]{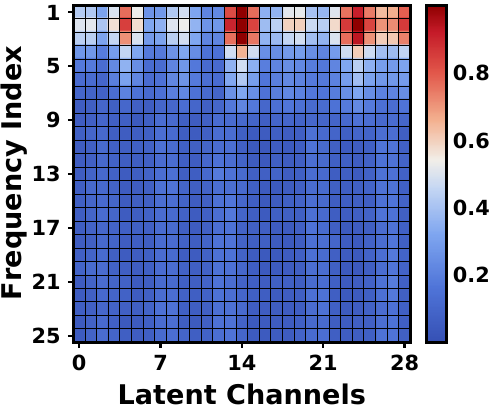}
	}%
 	\subfloat[NADEx]{
		\includegraphics[scale=0.45]{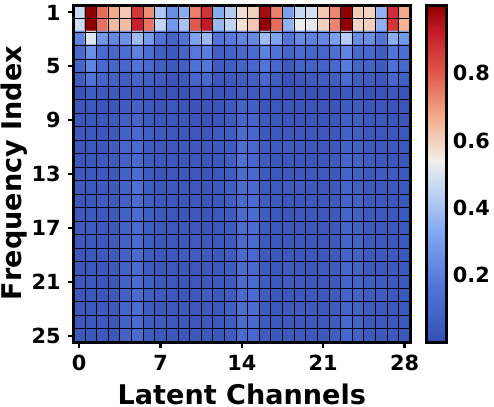}
	}%
	\centering

    \caption{Visualization of learned spectral energy distributions on ICEWS14. The y-axis represents frequency indices, and the x-axis represents latent channels. \textcolor{red}{Red} indicates higher spectral energy.}
	\label{app.Spec-viz}
\end{figure*}

\section{Experimental Analysis}

\subsection{Spectral Energy Distributions} \label{D.1}

Figure \ref{app.Spec-viz} shows that the four variants learn markedly different spectral energy distributions on ICEWS14. In FreqDiff w/o. $Freq$, the energy is highly concentrated in the lowest frequency band, while most frequency–channel positions remain weakly activated, indicating that the model mainly relies on dominant low-frequency temporal signals and lacks sufficient capacity to capture diverse evolutionary patterns. DiffuTKG and NADEx exhibit a similar tendency, where spectral responses are sparse and mostly confined to a few low-frequency regions, suggesting that conventional diffusion-based TKG extrapolation models still under-utilize frequency-domain information in historical event dynamics. In contrast, FreqDiff presents a more distributed and structured energy pattern across both frequency indices and latent channels, with visible activation not only in the low-frequency region but also in middle and higher frequency bands. This indicates that the proposed frequency-aware design enables the model to preserve richer spectral components, including stable long-term trends and more fluctuating short-term relational changes. Therefore, the visualization provides intuitive evidence that spectral modeling introduces complementary temporal signals beyond standard time-domain representations, which helps explain the improved extrapolation performance of FreqDiff.

\begin{figure}[t]
	\centering
    \hspace*{-0.3cm}%
	\subfloat[Impact of $\lambda$]{
		\begin{tabular}{@{}c@{\hspace{0.2cm}}c@{}}
			\includegraphics[scale=0.16]{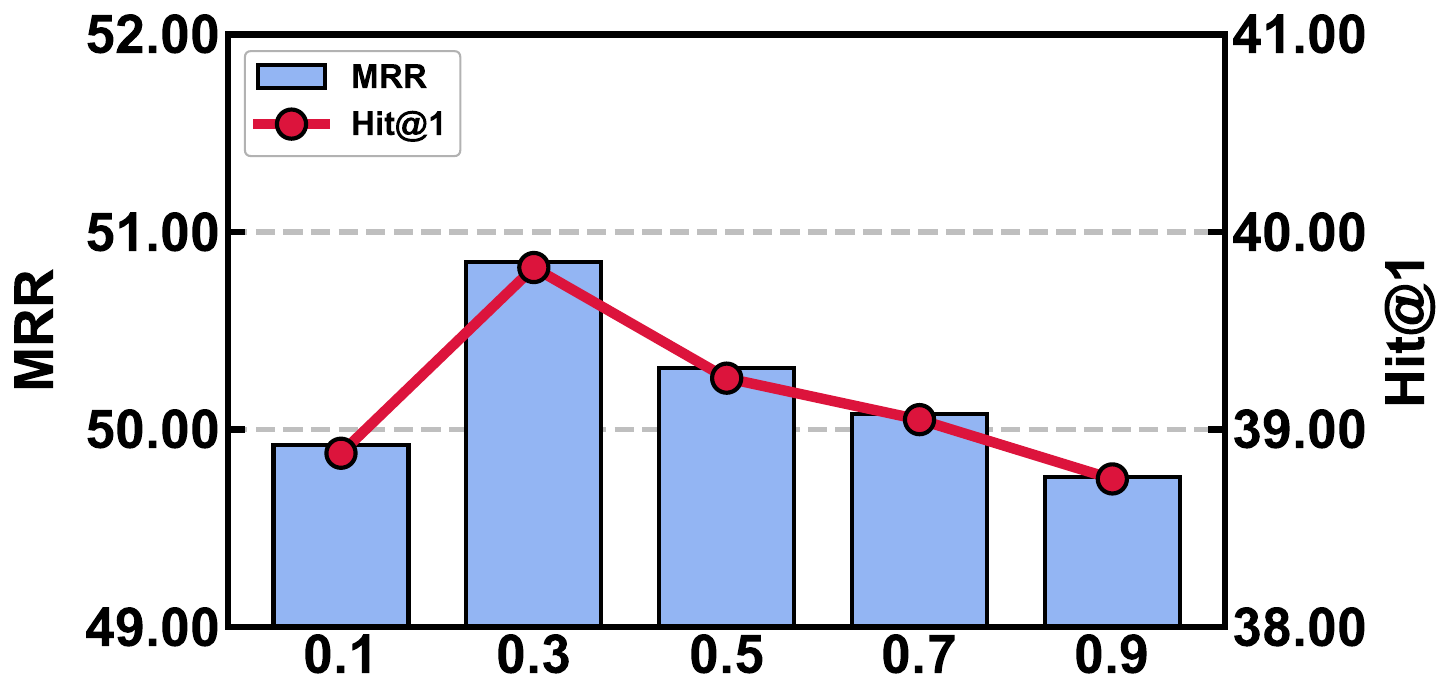} &
			\includegraphics[scale=0.16]{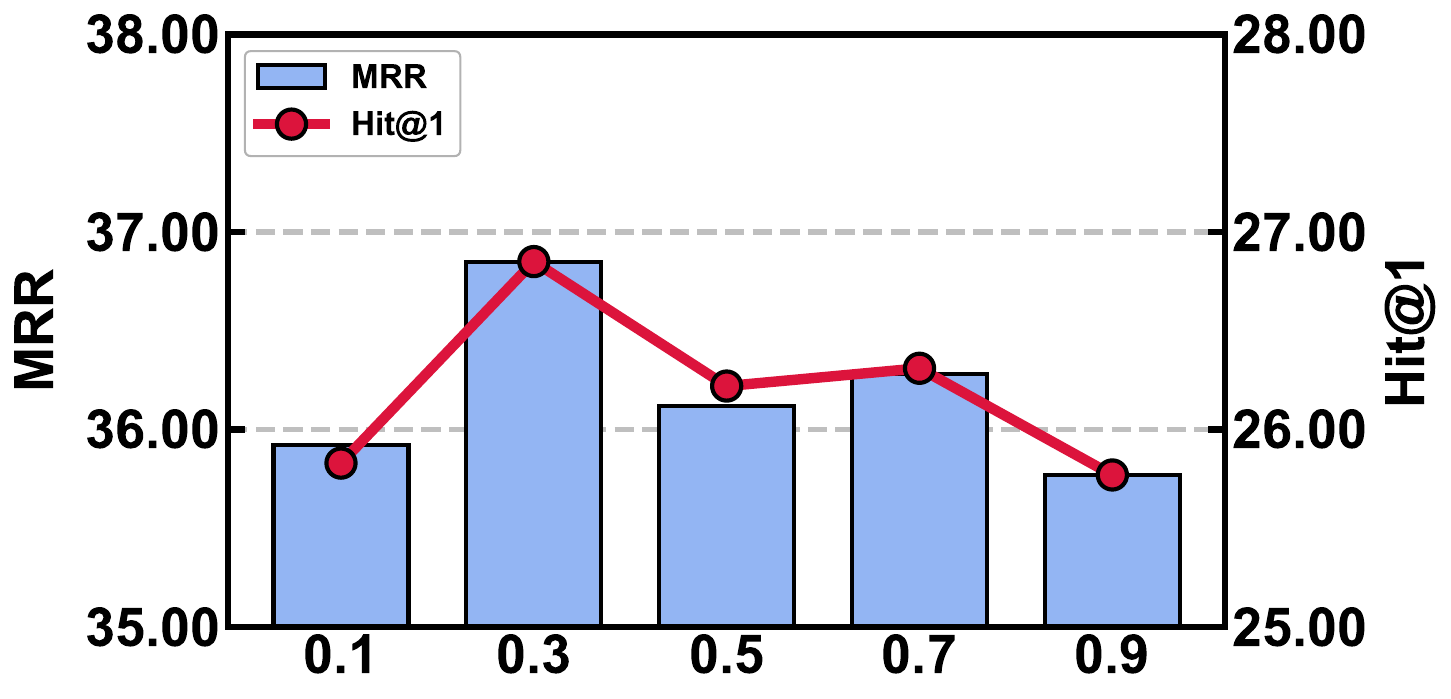}
		\end{tabular}
	}\\

    \hspace*{-0.3cm}%
	\subfloat[Impact of $\alpha$]{
		\begin{tabular}{@{}c@{\hspace{0.2cm}}c@{}}
			\includegraphics[scale=0.16]{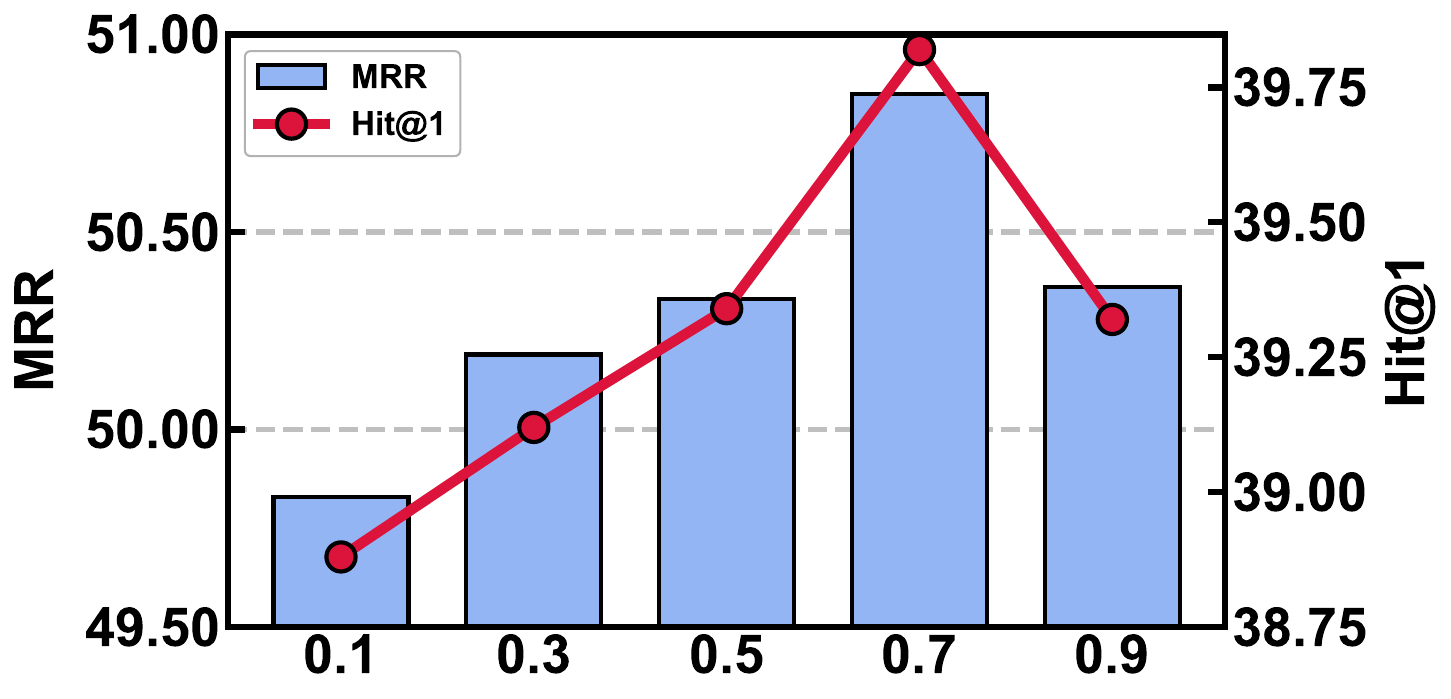} &
			\includegraphics[scale=0.16]{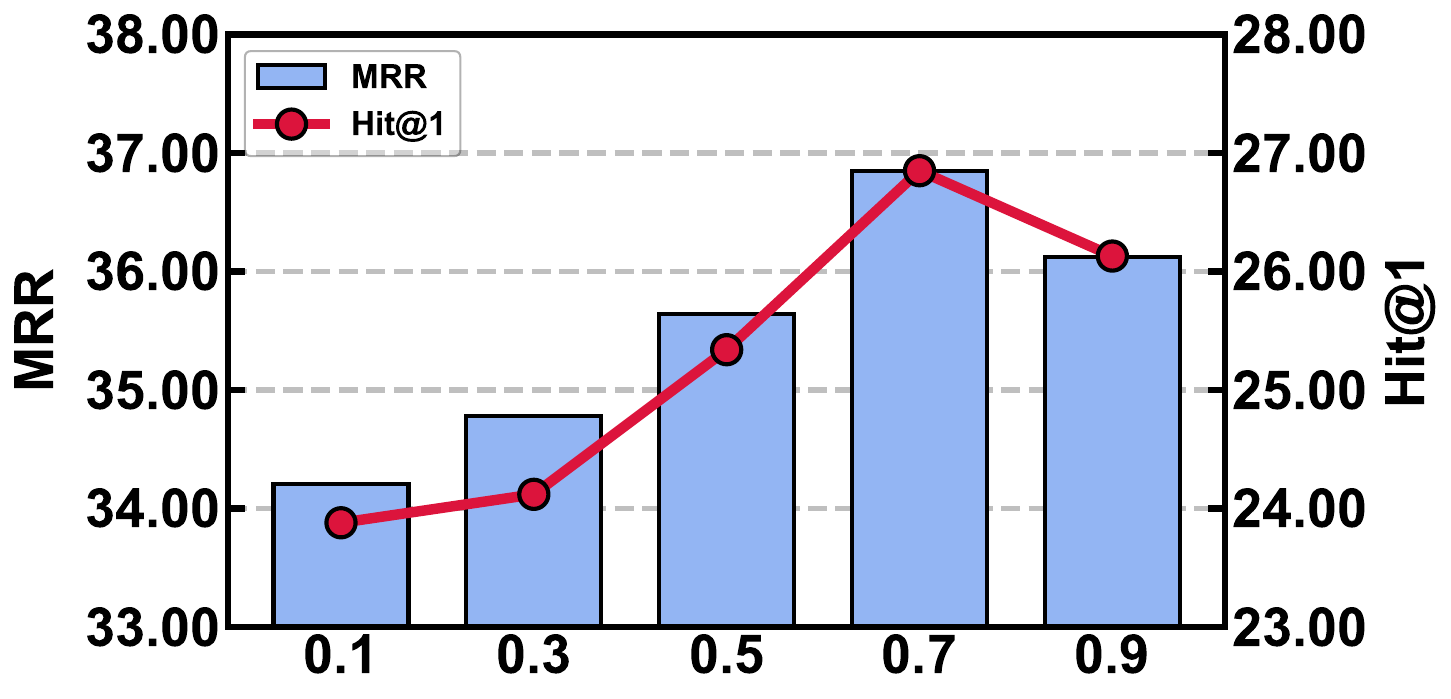}
		\end{tabular}
	}

	\caption{Hyper-parameter Sensitivity analysis on ICEWS14 (left) and ICEWS18 (right) datasets.}
	\label{app.Hyper}
\end{figure}

\subsection{Sensitivity Analysis}\label{D.3}

Table \ref{app.Hyper} evaluates the sensitivity of the weighting coefficient $\lambda$ on ICEWS14 and ICEWS18, where MRR and Hit@1 exhibit a generally consistent trend across the two datasets. When $\lambda$ increases from 0.1 to 0.3, both metrics improve noticeably, indicating that a moderate strength of the corresponding regularization/objective term can effectively enhance representation learning and improve future fact prediction. The best performance is obtained at $\lambda$=0.3 on both datasets, suggesting that this setting provides the most balanced contribution between the main prediction objective and the auxiliary constraint. However, when $\lambda$ continues to increase beyond 0.3, the performance gradually declines, especially on ICEWS14, where both MRR and Hit@1 show a clear downward trend from 0.5 to 0.9. This implies that an excessively large $\lambda$ may overemphasize the auxiliary learning signal and weaken the model’s ability to optimize the primary extrapolation objective. For the fusion coefficient $\alpha$, both datasets achieve strong performance around $\alpha$ = 0.7, showing that temporal modeling should remain dominant while spectral calibration provides complementary information.

\begin{table*}[t]
\setlength{\tabcolsep}{5pt} 
\fontsize{9pt}{10pt}\selectfont 
\caption{Case study of future entity prediction on ICEWS14. Given a query fact with the target entity masked, the table reports the top-5 predicted entities and their confidence scores produced by FreqDiff, FreqDiff w/o. $Freq$, and DiffuTKG. The red entries denote the ground-truth entities.}
\begin{tabular}{@{}cccc@{}}
\toprule
 & FreqDiff & \begin{tabular}[c]{@{}c@{}}FreqDiff\\ w/o. $Freq$\end{tabular} & DiffuTKG \\ \midrule
\begin{tabular}[c]{@{}c@{}}\textbf{Case \#1}\\ \textbf{Date}:\\ 2014-12-18\\ \textbf{Query}:\\ (Democratic Party (Nigeria),\\ \textit{Criticize or denounce}, ?)\\ \textbf{Gold}:\\ Muhammadu Buhari\end{tabular} & \begin{tabular}[c]{@{}c@{}}\textcolor{red}{1. Muhammadu Buhari}\\ (0.79056861)\\ 2. Citizen (Nigeria)\\ (0.06997431)\\ 3. Congress (Nigeria)\\ (0.04214321)\\ 4. Government (Nigeria)\\ (0.03278727)\\ 5. Head of Government (Nigeria)\\ (0.00737355)\end{tabular} & 
\begin{tabular}[c]{@{}c@{}}1. Citizen (Nigeria)\\ (0.712012)\\ \textcolor{red}{2. Muhammadu Buhari}\\ (0.145593)\\ 3. Government (Nigeria)\\ (0.028617)\\ 4. Congress (Nigeria)\\ (0.024223)\\ 5. Boko Haram\\ (0.011203)\end{tabular} & 
\begin{tabular}[c]{@{}c@{}}1. Citizen (Nigeria)\\ (0.231514)\\ 2. Government (Nigeria)\\ (0.184501)\\ \textcolor{red}{3. Muhammadu Buhari}\\ (0.076782)\\ 4. Chibuike Rotimi Amaechi\\ (0.029169)\\ 5. Ministry (Nigeria)\\ (0.029161)\end{tabular} \\ \midrule
\begin{tabular}[c]{@{}c@{}}\textbf{Case \#2}\\ \textbf{Date}:\\ 2014-12-25\\ \textbf{Query}:\\ (Ethiopia,\\ \textit{Sign formal agreement}, ?)\\ \textbf{Gold}:\\ Sudan\end{tabular} & 
\begin{tabular}[c]{@{}c@{}}\textcolor{red}{1. Sudan}\\ (0.77821444)\\ 2. China\\ (0.14515885)\\ 3. Portugal\\ (0.00858667)\\ 4. Angola\\ (0.00622532)\\ 5. France\\ (0.00528247)\end{tabular} & 
\begin{tabular}[c]{@{}c@{}}1. Portugal\\ (0.471355)\\ \textcolor{red}{2. Sudan}\\ (0.240765)\\ 3. China\\ (0.106766)\\ 4. United Arab Emirates\\ (0.070986)\\ 5. Angola\\ (0.012471)\end{tabular} & 
\begin{tabular}[c]{@{}c@{}}1. Portugal\\ (0.118397)\\ 2. China\\ (0.081131)\\ 3. Iran\\ (0.068646)\\ 4. South Africa\\ (0.065919)\\ 5. Ethiopia\\ (0.055098)\end{tabular} \\ \bottomrule
\end{tabular}
\label{case-study}
\end{table*}

\subsection{Case Study}

Table \ref{case-study} offers a concrete comparison of how different models rank candidate entities under realistic temporal extrapolation scenarios. In Case 1, for the query (Democratic Party (Nigeria), Criticize or denounce, ?), FreqDiff correctly predicts Muhammadu Buhari as the top-ranked entity with a confidence score of 0.7906, which is substantially higher than the scores assigned to other candidates. This shows that FreqDiff can identify the specific political figure most likely to be involved in the future event, rather than merely selecting broad and frequent entities such as Citizen (Nigeria) or Government (Nigeria). In contrast, FreqDiff w/o. Freq ranks the correct entity second, while DiffuTKG places it third with a much lower confidence score. This comparison suggests that removing frequency modeling weakens the model’s ability to capture discriminative temporal signals, causing it to rely more heavily on generic co-occurrence patterns. A similar observation can be made in Case 2. For the query (Ethiopia, Sign formal agreement, ?), FreqDiff ranks Sudan first with a confidence score of 0.7782, whereas FreqDiff w/o. Freq ranks it second and DiffuTKG fails to prioritize the correct answer, instead assigning higher ranks to countries such as Portugal, China, and Iran. These results indicate that the frequency-aware module helps FreqDiff capture relation-specific temporal regularities and distinguish the most contextually appropriate future entity from plausible but less accurate alternatives. Overall, the case study demonstrates that spectral information not only improves quantitative performance but also leads to more reliable and interpretable entity ranking in future fact prediction.

\begin{figure*}[t]
\centering

\begin{eventbox}{Muhammadu Buhari-linked input events}
\begin{lstlisting}[style=eventstyle]
[01] (Muhammadu Buhari, Give ultimatum, Democratic Party (Nigeria), 2014-04-18)
[02] (Muhammadu Buhari, Make statement, Democratic Party (Nigeria), 2014-10-16)
[03] (Muhammadu Buhari, Accuse, Democratic Party (Nigeria), 2014-11-20)
[04] (Democratic Party (Nigeria), Criticize or denounce, Muhammadu Buhari, 2014-12-04)
[05] (Democratic Party (Nigeria), Make statement, Muhammadu Buhari, 2014-12-04)
[06] (Democratic Party (Nigeria), Praise or endorse, Muhammadu Buhari, 2014-12-12)
[07] (Democratic Party (Nigeria), Criticize or denounce, Muhammadu Buhari, 2014-12-15)
[08] (Democratic Party (Nigeria), Criticize or denounce, Muhammadu Buhari, 2014-12-15)
[09] (Democratic Party (Nigeria), Criticize or denounce, Muhammadu Buhari, 2014-12-15)
[10] (Democratic Party (Nigeria), Accuse, Muhammadu Buhari, 2014-12-15)
\end{lstlisting}
\end{eventbox}

\vspace{4pt}

\begin{eventbox}{Citizen (Nigeria)-linked input events}
\begin{lstlisting}[style=eventstyle]
[01] (Democratic Party (Nigeria), Use conventional military force, Citizen (Nigeria), 2014-01-13)
[02] (Citizen (Nigeria), Reject, Democratic Party (Nigeria), 2014-01-24)
[03] (Citizen (Nigeria), Reduce relations, Democratic Party (Nigeria), 2014-02-08)
[04] (Democratic Party (Nigeria), Make an appeal or request, Citizen (Nigeria), 2014-02-10)
[05] (Democratic Party (Nigeria), Appeal for military protection or peacekeeping, Citizen (Nigeria), 2014-02-10)
[06] (Democratic Party (Nigeria), Make empathetic comment, Citizen (Nigeria), 2014-02-19)
[07] (Democratic Party (Nigeria), Criticize or denounce, Citizen (Nigeria), 2014-02-25)
[08] (Citizen (Nigeria), Reject, Democratic Party (Nigeria), 2014-03-07)
[09] (Democratic Party (Nigeria), Make pessimistic comment, Citizen (Nigeria), 2014-05-02)
[10] (Democratic Party (Nigeria), Demand, Citizen (Nigeria), 2014-06-18)
[11] (Citizen (Nigeria), Accuse, Democratic Party (Nigeria), 2014-07-28)
[12] (Citizen (Nigeria), Criticize or denounce, Democratic Party (Nigeria), 2014-08-01)
[13] (Democratic Party (Nigeria), Accuse, Citizen (Nigeria), 2014-09-15)
[14] (Democratic Party (Nigeria), Threaten, Citizen (Nigeria), 2014-09-15)
[15] (Democratic Party (Nigeria), Accuse, Citizen (Nigeria), 2014-10-09)
[16] (Democratic Party (Nigeria), Threaten, Citizen (Nigeria), 2014-10-13)
[17] (Democratic Party (Nigeria), Refuse to yield, Citizen (Nigeria), 2014-10-13)
[18] (Citizen (Nigeria), Reduce relations, Democratic Party (Nigeria), 2014-10-13)
[19] (Democratic Party (Nigeria), Bring lawsuit against, Citizen (Nigeria), 2014-10-14)
[20] (Democratic Party (Nigeria), Accuse, Citizen (Nigeria), 2014-10-16)
[21] (Citizen (Nigeria), Criticize or denounce, Democratic Party (Nigeria), 2014-10-20)
[22] (Citizen (Nigeria), Reduce relations, Democratic Party (Nigeria), 2014-10-31)
[23] (Citizen (Nigeria), Make optimistic comment, Democratic Party (Nigeria), 2014-11-14)
\end{lstlisting}
\end{eventbox}

\caption{Input event histories associated with the correct candidate and the misleading high-ranked candidate in Case \#1.}
\label{fig:case-history}

\end{figure*}

We further analyze the composition of the decoded subject histories to understand why different models favor different candidate entities. Taking Case \#1 as an example, Figure \ref{fig:case-history} shows that the two competing candidates are supported by different types of historical evidence. The entity \textit{Muhammadu Buhari}, which is ranked first by FreqDiff, is associated with only 10 input events. However, these events are highly query-relevant, as they involve direct interactions with the \textit{Democratic Party (Nigeria)} and include several recent criticism- and accusation-related events immediately before the query timestamp. In contrast, \textit{Citizen (Nigeria)}, which is selected as the top-1 prediction by FreqDiff w/o. $Freq$ and DiffuTKG, is linked to 23 input events. Although this candidate appears more frequently in the subject history, its associated events are more generic and broadly reflect interactions between the party and a collective political actor, rather than providing specific evidence for the masked entity in the given query. This comparison suggests that the baselines are more easily biased toward historically frequent candidates, especially when the entity has dense but less discriminative historical links. By contrast, FreqDiff ranks \textit{Muhammadu Buhari} first despite its fewer direct connections, indicating that the proposed spectral-aware filterbank helps to emphasize temporally informative and query-specific evidence rather than relying on raw historical frequency.

\end{document}